\documentclass[11pt,breaklinks,backref=page,colorlinks=true]{article}

\usepackage[preprint]{acl}

\usepackage{times}
\usepackage{latexsym}

\usepackage[T1]{fontenc}

\usepackage[utf8]{inputenc}

\usepackage{microtype}

\usepackage{inconsolata}

\usepackage{graphicx}
\usepackage{amssymb}
\usepackage{amsfonts}
\usepackage{algorithm}
\usepackage[noend]{algpseudocode}
\usepackage{markdown}
\usepackage{spverbatim}

\usepackage{url}
\usepackage{graphicx}
\usepackage{subcaption}
\usepackage{caption}
\usepackage{booktabs}

\renewcommand*{\backref}[1]{}
\renewcommand*{\backrefalt}[4]{%
  \ifcase #1
    Not cited.%
  \or
    Cited on page #2.%
  \else
    Cited on pages #2.%
  \fi
}

\usepackage{inconsolata} 
\usepackage[noabbrev]{cleveref}
\usepackage[group-separator={,}]{siunitx}
\crefname{section}{\S}{\S\S}
\Crefname{section}{\S}{\S\S}
\crefname{table}{Tab.}{}
\crefname{figure}{Fig.}{}
\crefname{algorithm}{Algorithm}{}
\crefname{equation}{eq.}{}
\crefname{appendix}{App.}{}
\crefname{thm}{Theorem}{}
\crefname{prop}{Proposition}{}
\crefname{cor}{Corollary}{}
\crefname{observation}{Observation}{}
\crefname{assumption}{Assumption}{}
\crefformat{section}{\S#2#1#3}
\usepackage{fontawesome5}

\usepackage{longtable}
\usepackage{array}

\usepackage[]{hyperref}

\usepackage[noabbrev]{cleveref}
\title{Exploring Sparse Autoencoders in \\ Text-Based Causal Confounding Adjustment}

\author{Mian Zhong\\
 Johns Hopkins University \\
 \texttt{mzhong8@jh.edu} \\
 \And
  \textbf{Katherine A. Keith}\\
  Williams College / Cohere \\
  \texttt{kak5@williams.edu} \\
  \And
 \textbf{Anjalie Field}\\
 Johns Hopkins University \\
 \texttt{anjalief@jhu.edu} 
\\
}

\begin{document}
\maketitle
\begin{abstract}
In many settings, studying causal questions based on text data requires adjusting for confounding information within texts. 
Yet there is a tradeoff in constructing text representations for adjustment: they must be sufficiently \emph{large and/or dense} to preserve the confounding variables necessary for unbiased effect estimation, but sufficiently \emph{small and/or sparse} to satisfy finite-sample overlap and yield low-variance estimates. To address this tradeoff, we turn to sparse autoencoders (SAEs), and propose a novel causal adjustment pipeline that iteratively selects a minimal set of SAE features via conditional independence tests.
We find that SAE representations achieve better adjustments (lower bias and and higher coverage) than alternative representations in standard semi-synthetic evaluations with binary confounders, and their interpretability offers opportunities for falsification.
We also introduce a more realistic semi-synthetic evaluation that uses multi-label data as the unobserved confounders and find off-the-shelf adjustment methods require increased investigation for these more complex settings. \faGithub\ Code: 
\href{https://github.com/mianzg/sae-text-confounder}{sae-text-confounder}

\end{abstract}

\section{Introduction}

Does the type of media coverage of a legislative bill affect the number of legislators who vote for it?
Does opting-in for an AI peer reviewer affect the acceptance of a manuscript to a venue? Does browsing on a mobile device versus computer affect the rates at which people share news articles? These are all questions about the \emph{causal effects} of a \emph{treatment} variable (e.g., media coverage type, review policy, or reading device) on an \emph{outcome} variable (e.g., number of votes, paper acceptance, or article popularity). Furthermore, these are the kinds of causal questions in which randomized controlled trails may be inaccessible or infeasible and researchers may only have access to observational data. In these observational settings, language data is often a confounding variable that must be adjusted for to obtain unbiased causal estimates. For example, the topic of a news article may influence a user's choice of device and proclivity to share.

Adjusting for confounding variables is well-studied in causal inference. 
However, adjusting for text-as-confounders remains difficult due to the high-dimensional nature of text and the limited interpretability of black-box text representations and causal adjustment methods \citep{keith-etal-2020-text,feder-etal-2022-causal}. Often, the challenges faced in applied work are mismatched with the ``best'' estimators. 
For example, while double machine learning is theoretically unbiased and can accommodate high-dimensional data types \citep{chernozhukov2017double}, the black-box nature of this method makes it difficult to detect failures such as ``lack of common support'' \citep{hill2013assessing} in applied settings. \citet{roberts2020adjusting} propose combining structural topic models with causal matching to increase interpretability, reduce extrapolation, and allow human oversight of adjustment quality. Yet, they rely on the topic model representations from bag-of-words input sufficiently capturing confounders. 

Recent work has demonstrated that sparse autoencoders~(SAEs) \citep{makhzani2013k} produce atomic, expressive, and interpretable features in the decomposition of large language models \citep{Huben2024,bricken2023towards,templeton2024scaling,Bills2023}.
Two characteristics of SAEs suggest their potential value for causal adjustment. First, SAEs have been shown to be expressive and interpretable representations in \emph{correlational} analyses of text data, including in comparison to topic modeling approaches \citep{movva2025sparse,jiang2025towards,Choi2025}.
Second, the sparsity of an SAE representation is an asset when adjusting for confounding since as the dimension of the adjustment set grows, strict overlap decreases \citep{d2021overlap} and estimator variance increases. Because only a small number of SAE features are active for a given text, the effective adjustment set is small even when the number of SAE features are large.

Motivated by these charateristics, we propose, implement, and empirically evaluate a causal adjustment pipeline with SAE representations.
Given standard assumptions in observational causal inference (overlap, homogeneous effect etc.) and text-specific assumptions such as \emph{causally sufficient representations} (see \cref{identification-assumptions} for specifics), our pipeline aims to achieve unbiased, low-variance estimates along with interpretable diagnostics supporting falsification for applied researchers. At a high-level, our pipeline trains an SAE on all input text and then uses a Lasso logistic regression model with conditional independence tests to select a minimal subset of SAE features that correlate with treatment. 
Then our subsetted SAE features can be used with downstream causal estimators; in this work, we compare double machine learning and matching.
We empirically compare our method to alternatives using four semi-synthetic evaluations, including new multi-class and multi-label confounder evaluations.\footnote{In the causal literature, \emph{semi-synthetic} evaluations use some variables from real-world data, i.e., confounders and text data, and other variables are created synthetically so the true causal effect is known and can be evaluated; see \citet{keithrct} for a comparison of other causal evaluation strategies.} In evaluations with binary confounders, with similar setups as prior work~\citep{roberts2020adjusting, veitch2020adapting, veljanovski-wood-doughty-2024-doublelingo}, we find
SAEs recover simulated treatment effects more accurately than alternatives.

Our primary contributions include: 

\begin{itemize}
    \item A method to combine SAEs with confounding adjustment methods that offers strong empirical performance in standard binary and multi-class semi-synthetic evaluation settings and diagnostics supporting falsification. 
    \item A new semi-synthetic evaluation for text as confounding, the first to our knowledge, that uses multi-label oracle confounding, moving closer to realistic text as confounding settings. Our results for this setting are mixed, encouraging applied researchers to be cautious when using black-box adjustment methods.
\end{itemize}

\section{Related Work}

\paragraph{Causal Inference with Text}
Applications of causal inference with text span a variety of domains, such as linguistics, social sciences, and public health \citep{keith-etal-2020-text,feder-etal-2022-causal}. While causal adjustment methods often involve using standard off-the-shelf approaches like propensity matching or regression with extracted text features \citep{keith-etal-2020-text}, a few studies have designed methods for text specifically. Approaches involve text classifiers \citep{wood-doughty-etal-2018-challenges}, fine-tuning pre-trained models with causal-focused loss objectives \citep{veitch2020adapting}, clustering off-the-shelf text embeddings \citep{zhang-etal-2023-causal-matching}, directly applying pre-trained embeddings with double machine learning~\citep{schulte2025adjustment}, combining low-rank training of large language models with double machine learning \citep{veljanovski-wood-doughty-2024-doublelingo}, or using zero-shot LLMs in proximal causal inference \citep{chen2024proximal}. 
Similar to our motivation, \citet{roberts2020adjusting} propose matching representations from topic models, with the motivation that interpretable representations and adjustments methods are particularly useful in text settings, as text is human-readable, facilitating human oversight of the analysis. We follow this motivation in focusing on interpretable representations from SAEs.

\paragraph{Sparse Autoencoders (SAEs)}
SAEs are trained to compress data into a hidden representation and then reconstruct the original input. They differ from general autoencoders in that they enforce sparsity in the hidden representation. For example, K-Sparse autoencoders \citep{makhzani2013k} enforce that at most $k$ dimensions in the hidden representations are non-zero. 
SAEs have gained popularity for enabling interpretation of LLMs, where constructing sparse high-dimensional representations of model activation vectors aims to provide visibility into what information is encoded where \citep{bricken2023towards,Bills2023,Huben2024,templeton2024scaling,gao2025scaling}.
 Recent works have continued to demonstrate their potential as direct representations of text data that can facilitate discovering concepts in datasets \citep{jiang2025towards,Choi2025,peng2025use} and hypothesis generation \citep{movva2025sparse}. Our work similarly leverages SAEs as representations of text, but for adjusting for confounders in causal inference studies.

\section{Preliminaries on Confounding Adjustment}
Understanding causal effects with (large-scale) observational data requires statistically adjusting for confounders on the ``backdoor path'' between the treatment $T$ and the outcome $Y$.
We limit our scope to domains in which we have access to texts $D =\{d_1, \dots, d_N\}$ that encode unobserved confounder(s) $U$ (e.g., the semantic content of a bill). 
As in most modern text-based work, we must obtain a representation of $D$ which we denote $X \in \mathbb{R}^{N\cdot M}$. 

\subsection{Causal identification assumptions} \label{identification-assumptions}

As in all causal studies, our method and causal estimates rest on causal identification assumptions that cannot be verified from data alone, including (1) No unmeasured confounding, (2) Overlap/Common Support, (3) Consistency, and (4) Homogeneous effects.\footnote{See \cref{app:identification-assumptions} for a refresher.}
 We also make the two identification assumptions specific to text:
\begin{enumerate}
    \item \emph{Pre-treatment text.} All texts are generated temporally prior to treatment.
    \item \emph{Causally sufficient representations.} We can infer from text ``representations that preserve sufficient information for causal identification'' \citep{veitch2020adapting}.
\end{enumerate}

Notably, (2) can be violated in several ways. First, the text may not carry all confounding information, meaning no choice of representation, SAE or otherwise, is sufficient.
Other violations are subtler and ones we explore in this work: for example
the dimensionality of an inferred representation may be too low to represent all confounding information.
This exposes a tension between sufficiency and sparsity. Causal sufficiency pushes toward richer representations, since any confounding the representation fails to encode results in a biased effect estimate. Yet, as the dimensionality or density of the adjustment set grows, strict overlap becomes increasingly difficult to satisfy \citep{d2021overlap} and the variance of the resulting estimates grows. We hypothesize SAEs may be one representation (of possibly many) that strikes a balance between sufficiency and sparsity.

\subsection{Adjustment estimators} 

Once we obtain text representations $X$, they can be plugged into downstream causal adjustment estimators. In this work, we compare double machine learning~(DoubleML) \citep{chernozhukov2017double} and matching \citep{Stuart_2010}.
DoubleML has gained popularity due to its asymptotic unbiasedness\footnote{Under the causal identification assumptions of \cref{identification-assumptions}, standard regularity conditions, and the requirement that the nuisance estimators converge at rate $o(n^{-1/4})$; cross-fitting removes the bias from overfitting the nuisance models.} and its flexibility to combine machine learning models for the treatment and outcome processes. However, as we discuss later in \cref{subsec:eurlex}, the black-box nature of DoubleML can also hide finite data issues such as ``lack of common support''. Matching estimators instead find ``matched'' treatment and control units in the representation space, such that the covariate distributions are balanced across the two groups. We hypothesize that matching will benefit from the sparsity of SAE representations because matching quality depends on distance in the adjustment space, and distances degrade rapidly as dimension grows: under a dense high-dimensional representation, all pairs become roughly equidistant and the nearest available control may still be a poor match.
Additionally, because SAE features are individually interpretable, a match can be inspected directly — an analyst can read off which features the matched pair shares and which it does not, rather than assessing balance only through aggregate summary statistics.

\section{Pipeline: Causal Adjustment with SAEs}\label{sec:pipeline}
We propose a pipeline that trains texts $D$ into SAE features and then uses statistical testing to obtain a minimal set of these features that are correlated with treatment in order to minimize variance.

\paragraph{Step 1: Infer Text Representations via Sparse Autoencoders}
Given $N$ input text samples, we first encode input texts using off-the-shelf pre-trained text embeddings $D_{\text{embed}}:=\{e_{1},\dots, e_N\}, e_i \in \mathbb{R}^E$. Then all embeddings are used to train a top-K sparse autoencoder with dimension $M$~\citep{makhzani2013k}: 
\[h_i = \text{TopK}(W\cdot e_i + \text{bias}), W\in \mathbb{R}^{M\cdot E}\]
 that enforces sparsity by keeping K active neurons out of M neurons.
We thus obtain SAE representation $X_{SAE}\in\mathbb{R}^{N\cdot M}$. We discuss the sensitivity to these hyperparameters in \cref{subec:hyperparams}.

\paragraph{Step 2: Subset to Minimal SAE Features that Correlate with Treatment}
Many different subsets of features can be ``valid'' adjustment sets that block the backdoor path between treatment and outcome and result in unbiased effect estimates. Choosing a minimal adjustment set results in lower variance in the estimates, and for finite data samples, fewer variables means its easier to satisfy strict overlap resulting in inference that is actually tractable \citep{d2021overlap,tamarchenko2023combining}.
Thus, rather than conducting adjustment on the full $X_{SAE}$, we first introduce a process to select the SAE features that are most informative on treatment $T$,\footnote{We acknowledge some work selects features that correlate with $Y$ to reduce variance; we leave exploration of these differences to future work.} leveraging the atomic and interpretable tendencies of SAE dimensions.
To achieve this, data is split into a train and a test set, 
and we use a logistic regression model with Lasso regularization on the train set to select features on $T$. 
The selection follows a grid search style: Starting with no features, we iteratively decrease regularization strength to select a subset of SAE features $X'\in\mathbb{R}^{N\cdot M^{'}}, M^{'}\leq M$. 
We then perform the following validation on the test set and stop at a desirable minimal set $X^{\star}$.

\paragraph{Step 3: Optimize Feature Subsetting with 
Conditional Independence Tests}
A minimal sufficient set $X^{*}$ is a subset such that the treatment variable is conditionally independent of the remaining features in $X$ given the selected subset $X^{\star}$:
\[T \perp X \setminus X^{\star}\mid X^{\star}\]
To validate if $X'$ from Step 2 satisfies such conditional independence, we employ likelihood ratio test between two logistic regression models on $T$ with the subset $X'$ and the full set $X_{SAE}$ using the test set data. 
Loosely, the null hypothesis tests on whether using this subset $X'$ to model $T$ is equally good as using the full set $X_{SAE}$. Rejecting the null hypothesis implies that including more features from $X_{SAE}\setminus X'$ may provide more information about $T$. Thus, $T$ and $X\setminus X'$ are not yet conditionally independent.
Practically, we perform a chi-square test on the log-likelihood ratio with a significance level (e.g., 0.05). 
If the p-value is below the significance level, we return to Step 2 to include more features.

Finally, by iteratively running Step 2 \& 3, we reach a p-value equal to or larger than the significance level and obtain the desirable set of features
$X^{\star}\in\mathbb{R}^{N\cdot M^{\star}}, M^{\star} \leq M$. Note, it is possible that $X^{\star} = X_{SAE}$ the full set of features.

\paragraph{Step 4: Plug-In Final Subsetted Features to a Causal Estimator}
We complete adjustment using two methods: Coarsened Exact Match (CEM)~\cite{iacus2012, Stuart_2010} and double machine learning~(DoubleML) \citep{chernozhukov2017double}.
CEM converts each continuous-valued covariate into discrete bins and performs exact match on the discretized-valued samples. A bin is retained only if it contains both treatment and control samples.Then, causal estimands are calcualted over the retained bins.
The binning strategy can be fixed numeric cut-off, like topic mass $\geq 0.1$, or more adaptive rules, like binning by $(25\%, 50\%, 75\%)$ quantiles. 

Double ML integrates machine learning models for estimating nuisance functions. In our experiments, the propensity score model $e(X) = P(T = 1|X)$ is estimated using logistic regression with L2 penalty, and, the outcome regression $m_{t}(X) = E(Y|T=t, X)$ is estimated using linear regression. Both nuisance estimates are then combined to estimate the treatment effect. For both CEM and DoubleML we target the average treatment effect on the treated (ATT).

\subsection{Interpretable Falsification Diagnostics}
\label{sec:methods_falsification}

We identify two aspects of our pipeline that offer opportunities for falsification, i.e., identifying that the pipeline as-implemented failed to achieve its intended behavior or violated an assumption.

\paragraph{Selecting M} First, as discussed in \S\ref{identification-assumptions}, the SAE representation needs to be rich enough to capture confounding information. Our selection process in Step 2 and 3 offers a simple diagnostic to test this: if selection identifies $X^{\star} = X_{SAE}$, meaning every SAE dimension is informative of $T$, and $X_{SAE}$ does not capture any additional information, the choice of $M$ may be too small. As $M^{\star} < M$ does not guarantee that $X_{SAE}$ is sufficient, this heuristics offers an opportunity for falsification, rather than a test of correctness.

\paragraph{Qualitative Analysis of SAE features}

Second, following \citet{movva2025sparse}, we can map SAE dimensions to interpretable concepts by identifying commonalities in texts that have each neuron active. This interpretation allows a researcher to identify what specific aspects of the text were identified as reflecting confounders and check for face validity, i.e.,~if the identified confounders make intuitive sense based on their knowledge of the data. The presence of non-intuitive confounders or the absence of expected confounders would provide an opportunity to falsify the pipeline.

\section{Semi-Synthetic Experiments}

Semi-synthetic data -- using real covariates but a synthetic data generating process so true causal effects are known -- is used extensively in causal estimation research. Next, we describe our semi-synthetic set-up and the necessity of creating a new, more complex setting compared to previous work. 

\subsection{Data generating processes~(DGPs)}\label{subsec:dgp}
With access to texts and pre-defined true confounders (e.g., topics), we simulate treatment and outcome variables through DGPs. Then, adjustment approaches and causal estimators use text and synthetic treatment/outcomes during evaluation with no access to the true confounders. In our experiment, we represent a true confounder $U$ as a one-hot vector, for example, $U = [0, 1, 0]$ in a 3-class setting. Similar to prior work, we construct simulations where $U$ is single-dimensional, representing a single binary confounder~\citep{roberts2020adjusting, veitch2020adapting, veljanovski-wood-doughty-2024-doublelingo}, and we introduce new simulations in which $U$ reflects multi-class and multi-labeled documents. The DGP synthetically generates a binary treatment variable $T$ and a continuous outcome variable $Y$ for all simulations as follows:

\[T \sim \text{Bernoulli}(\sigma(\alpha +\beta U))\]
where $\sigma$ is the sigmoid function, $\alpha$ indicates a baseline strength to receive treatment, and a vector $\beta$ denotes confounding strengths for treatment. Then,

\[Y = \tau T + \gamma U + \epsilon, \epsilon \sim \mathcal{N}\] 
where a constant scalar $\tau$ indicates true treatment effect, $\gamma$ is a vector controlling confounding strength, and $\epsilon$ is a gaussian noise. \textbf{For all experiments, we set the true treatment effect $\tau = 2$} and noise to be $\epsilon \sim \mathcal{N}(0, 0.09)$. \cref{tab:confounding-parameters} specifies DGP parameters used in our experiments. 
\begin{table}[!h]
    \centering
    \resizebox{\linewidth}{!}{
    \begin{tabular}{lrrr}
        U & $\alpha$ & $\beta$ & $\gamma$ \\
        \midrule
        binary & $0$ & $[0,2]$ & $[0, 0.5]$\\
        multi-class & $0.5$ &$[0, 0.5, 1]$ & $[0.5, 1.0, 2.0]$ \\
        multi-labeled & $0.5$ & $[0.0, 0.1, 0.2, 0.3, 0.4]$ & $[0.5, 0.75, 1.0, 1.5, 2.0]$ \\
    \end{tabular}
    }
    \caption{Parameter set-ups for baseline strength $\alpha$ of receiving treatment, confounding strength $\beta$ for treatment, and $\gamma$ for outcome in our DGPs.}
    \label{tab:confounding-parameters}
\end{table}

\subsection{Data}
We use two datasets in our semi-synthetic DGPs. Label prevalence of both datasets and an example of treatment assignment distribution for different confounding settings can be found in \cref{app:data}.

\paragraph{20NewsGroups (20NG)} This dataset comprises 18,331 singly-labeled newsgroup posts~\citep{Lang95}. We manually merge original 20 labels with similar topics into 10 final labels. 
In our semi-synthetic simulation, we pick a more frequent label (i.e., computer) and a less frequent one (i.e., religion) as the true confounder to simulate binary confounder. We also craft a multi-class setting with 3 classes: computer, religion, and other.

\paragraph{EURLEX} EURLEX~\citep{chalkidis-etal-2021-multieurlex} is a multi-labeled multi-lingual dataset on EU laws. We use the English set and select the 5 most prevalent labels to set up the multi-labeled confounding, i.e., for each sample, we remove labels that are not in these 5 labels. Our final dataset contains 62,007 samples.
To our best knowledge, this is the first semi-synthetic simulation building on multi-labeled confounders.

\begin{table*}[ht]
\centering

\resizebox{0.95\linewidth}{!}{
\begin{tabular}{llrrrrrrrrrr}
\toprule
      U & Repr. & \multicolumn{4}{c}{\textbf{CEM}} & &\multicolumn{3}{c}{\textbf{DoubleML}} \\
     \midrule
      &  & \textbf{Bias}~$\downarrow$ & \textbf{RMSE}~$\downarrow$ & \textbf{Cov.}~\%~$\uparrow$ &\textbf{T-ret.\%}~$\uparrow$ & & \textbf{Bias}~$\downarrow$ & \textbf{RMSE}~$\downarrow$ & \textbf{Cov.}~\%~$\uparrow$\\
     \cline{3-6} \cline{8-10}\noalign{\vskip 2pt}
     
      Binary  & Unadjusted & 0.1530 & 0.1530 & 0.00 & \textnormal{n/a}\\
      (computer) & TIRM~(STM) & 0.1149 & 0.1163 & 0.00 & 2.20 & & 0.0663 & 0.0663 & 0.00\\
      & Embed & \textnormal{n/a} & \textnormal{n/a} & \textnormal{n/a} & \textnormal{n/a} & & 0.1873 & 0.1874 & 0.00 \\
      & SAE & 0.0361 & 0.0485 & \textbf{91.00} & 0.29 & & \textbf{0.0480} & \textbf{0.0480} & 0.00 \\
      & SAE$_{select}$ & \textbf{0.0312} & \textbf{0.0416} & 65.00 & \textbf{15.25} & & 0.0602 & 0.0604 & 0.00 \\ 
      \midrule
      
      Binary & Unadjusted & 0.0865 & 0.0865 & 0.00 & \textnormal{n/a} &  \\
    (religion) & TIRM~(STM) & 0.0466 & 0.0481 & 5.00 & 2.39 & & 0.0663 & 0.0663 & 0.00\\
     & Embed & \textnormal{n/a} & \textnormal{n/a} & \textnormal{n/a} & \textnormal{n/a} &  & 0.1200 & 0.1201 & 0.00  \\
     & SAE  & \textbf{0.0062} & \textbf{0.0246} & \textbf{92.00} & 0.40 & & \textbf{0.0291} & \textbf{0.0291} & 0.00\\
     & SAE$_{select}$ & 0.0390 & 0.0433 & 70.00 & \textbf{81.56} & & 0.0377 & 0.0377 & 0.00\\
     \midrule
     
    Multi-class & Unadjusted & 0.1572 & 0.1574 & 0.00  & \textnormal{n/a} \\
     & TIRM~(STM) & 0.1036 & 0.1112 & 28.00 & 2.14 & & 0.0801 & 0.0803 & 0.00\\
     & Embed & \textnormal{n/a} & \textnormal{n/a} & \textnormal{n/a} & \textnormal{n/a} & & 0.1906 & 0.1908 & 0.00 \\
     & SAE & \textbf{0.0205} & 0.0450 & \textbf{95.00} & 0.30 & & \textbf{0.0536} & \textbf{0.0536} & 0.00\\
     & SAE$_{select}$ & 0.0267 & \textbf{0.0444} & 77.00 & \textbf{4.77} & & 0.0579 & 0.0581 & 0.00\\

  \bottomrule
   \end{tabular}}
\caption{\textbf{Main results table for 20NG} with metrics for which lower ($\downarrow$) or higher ($\uparrow$) is better. We compare our method of using SAE representations ($M=128, K=32$) without and with selection (\emph{select}) with TIRM and pre-trained embeddings~(Embed), using a coarsened exact matching (CEM) estimator (cut-points at 0 and at 50\% quantile of positive value) and a Double ML estimator, with unadjusted estimate reported. For CEM, the percentage of treated retained~(T-ret.\%) after matching is reported. Metrics are the mean value from $100$ simulations.
}
\label{tab:pipeline-comparison-20ng}
\end{table*}

\subsection{Baseline Pipelines}
To compare with our framework using SAEs, we select two prior end-to-end methods: Topical Inverse Regression Matching~(TIRM)~\citep{roberts2020adjusting} and DoubleML using pre-trained text embeddings~\citep{schulte2025adjustment}.

\paragraph{TIRM}
Input texts $D$ are formed as bag-of-words using common pre-processing (e.g., remove stop words, numbers, punctuations, low-frequency words).
Then TIRM trains a structural topic model (STM)~\citep{roberts2013structural} with the treatment variable as a content covariate and conducts a refitting step to finalize text representations.
TIRM applies CEM matching on the constructed STM representations to conduct adjustment.
In our experiments, we set the SAE and STM representations to have the same number of dimensions for fair comparison. To isolate differences between representations and adjustment methods, we also report results when using STM representations with the same Double ML causal estimator as SAEs.

\paragraph{Off-the-shelf Embeddings + DoubleML}
Following \citet{schulte2025adjustment}, we use pre-trained text embeddings as covariates for the DoubleML estimator. For a fair empirical comparison, these same pre-trained embeddings are used in the first step of training the SAE. We do not use CEM with pre-trained embeddings due to their high-dimensionality as well as lack of evidence that individual dimensions are semantically meaningful.

\subsection{Metrics}
For each simulation, we obtain an estimated treatment effect $\hat{\tau}$ via ATT and its confidence interval $CI$. We compute mean bias, RMSE, and coverage across the simulations. Coverage is the percentage of CI's that covers the true treatment effect.
To analyze the selection step of the pipeline under matching, we evaluate overlap with effective sample size~(ESS) where a higher ESS suggests better overlap. Balance checking of matched data is measured with the absolute standardized mean difference (|SMD|) on the true confounder $U$. Overlap and balance diagnostics are important for evaluating the quality and credibility of matching adjustment. See \cref{sec:appendix:overlap} for detailed calculations.

\section{Results}
For each of the four semi-synthetic DGPs (\cref{subsec:dgp}), we run simulations with 100 different random seeds and compare our method to the baselines under both CEM and DoubleML causal estimators when available. We train SAE and TIRM with a model dimension $M = 128$ and set $K=32$ for SAE. We provide implementation details in \cref{app:experiment}.

\begin{figure*}[ht]
    \centering
    \includegraphics[width=\linewidth]{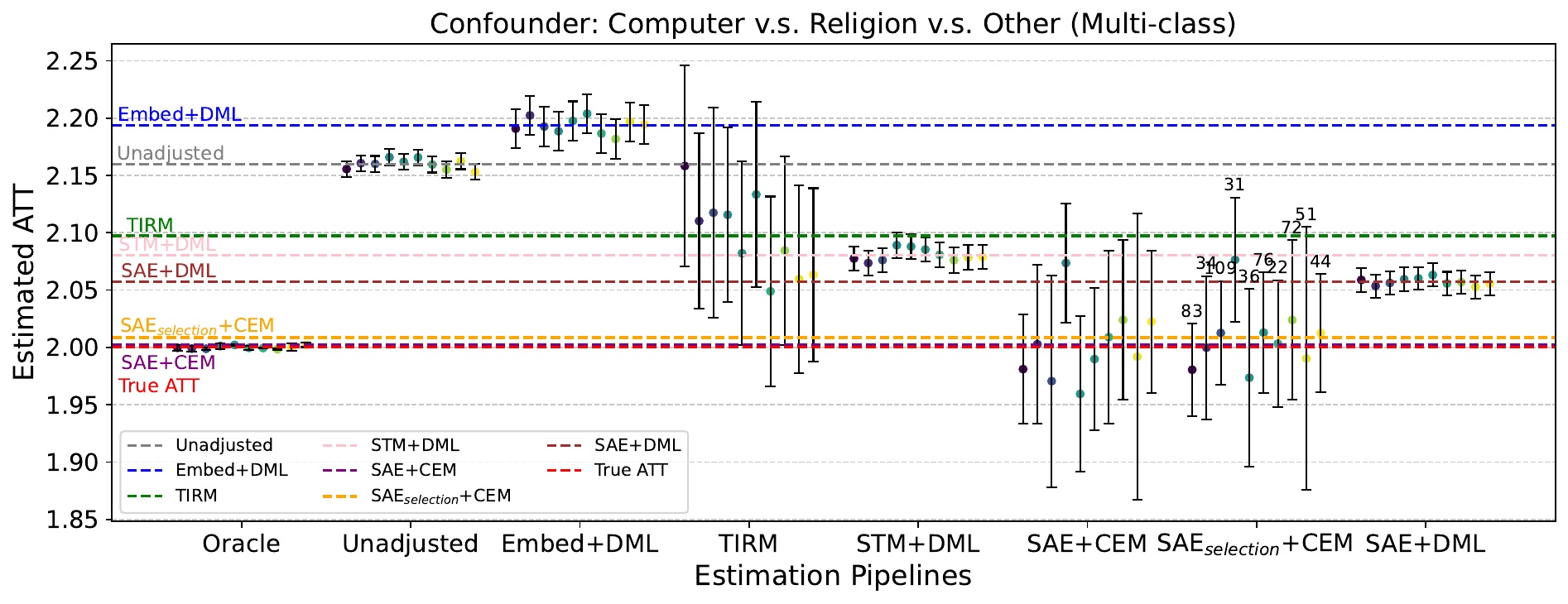}
    \caption{\textbf{Multi-class breakdown}. We report simulations across 10 (out of 100) randomly selected seeds for the oracle multi-class confounder ($U$) of computer versus religion versus other using the 20NewsGroups data. Horizontal lines are the average estimate across these 10 simulations. DML denotes DoubleML estimator. In SAE$_{selection}$+CEM, the number of selected features is listed on the top of the CI bars. We omit SAE$_{selection}$+DML as its performance not substantially different from SAE+DML as discussed from \cref{tab:pipeline-comparison-20ng}.
    }
    \label{fig:multiclass-m128}
\end{figure*}

\begin{table*}[ht]
    \centering
    \resizebox{0.95\linewidth}{!}{
    \begin{tabular}{llrrrrrrrrrc}
\toprule
       U & Repr. & \multicolumn{4}{c}{\textbf{CEM}} & &\multicolumn{3}{c}{\textbf{DoubleML}} \\
     \midrule
      &  & \textbf{Bias}~$\downarrow$ & \textbf{RMSE}~$\downarrow$ & \textbf{Cov.}~\%~$\uparrow$ &\textbf{T-ret.\%}~$\uparrow$ & & \textbf{Bias}~$\downarrow$ & \textbf{RMSE}~$\downarrow$ & \textbf{Cov.}~\%~$\uparrow$\\
     \cline{3-6} \cline{8-10}\noalign{\vskip 2pt}
     
    EURLEX & Unadjusted & 0.2863 & 0.2865 & 0.00 &\textnormal{n/a}\\
      multi-label & TIRM & 0.0335 & 0.0352 & 59.00 & \textbf{20.90} & & 0.1461 & 0.1464 &  0.00 \\
      & Embed & \textnormal{n/a} & \textnormal{n/a} & \textnormal{n/a} &  \textnormal{n/a}& & 0.2912 & 0.2914 & 0.00 \\
      & SAE & 0.0344 & 0.0638 & \textbf{92.00} & 0.25 & & \textbf{0.1324} & \textbf{0.1326} & 0.00 \\
      & SAE$_{select}$ & \textbf{0.0306} & \textbf{0.0558} & 91.00 & 0.42 & & 0.1336 & 0.1338 & 0.00 \\
      \bottomrule
    \end{tabular}}
    \caption{\textbf{Main results table for EURLEX} with metrics for which lower ($\downarrow$) or higher ($\uparrow$) is better. Refer to \cref{tab:pipeline-comparison-20ng} for notations and full set-ups.}
    \label{tab:pipeline-eurlex}
\end{table*}
\subsection{Results for binary and multi-class  (20NG)}\label{sec:results_20ng}

\cref{tab:pipeline-comparison-20ng} shows the results of all pipelines for the binary and multi-class settings with 20NewsGroups (20NG). SAE-based approaches achieve the lowest bias, lowest RMSE, and highest coverage across all three settings. 
Notably, for the multi-class setting, the bias for SAE+CEM (0.0205) and SAE$_{select}$+CEM (0.0267) is substantially lower than the unadjusted estimate (0.1572). Comparing to the two end-to-end baselines from prior work TIRM (0.1036)\footnote{TIRM includes CEM as the causal estimator so we omit the notation of ``CEM'' for it throughout the results section and refer TIRM+DoubleML to solely using STM representations with DoubleML estimator.} and Embed+DoubleML (0.1906), SAEs achieve better performance under both CEM and DoubleML, with or without selection (bias ranges from 0.0205 to 0.0579).

We hypothesize the decreased performance of TIRM is due to the dense nature of STM representations, leading to far fewer treated~(and control) units retained after matching (T-ret.\%).
The decreased performance of TIRM+DoubleML as compared to SAE+DoubleML additionally suggests that SAEs may better capture the topical information in $U$.
As the same dense embeddings ultimately underlie Embed+DoubleML and SAE+DoubleML, the performance differences here are more striking. We suspect that the sparse atomic nature of SAEs represent the topical information in $U$ in a way that is easier for simple off-the-shelf nuisance functions to learn, and more careful selection and tuning of nuisance functions for high-dimension text data, e.g., as explored in \citet{veljanovski-wood-doughty-2024-doublelingo}, may lead to more comparable performance between Embed+DoubleML and SAE+DoubleML. However, reducing the need for careful selection of nuisance functions and their hyperparameters offers easier implementations in practical settings.

Finally, in \cref{fig:multiclass-m128}, we display 10 out of the 100 random seeds for each pipeline. Notably, although DoubleML often achieves low bias, estimates also have much narrower confidence intervals, resulting in the zero coverage observed in \cref{tab:pipeline-comparison-20ng}. Additionally, there is a large variance in the selection step: ranging from 22 to 109 of the 128 SAE features selected. We hypothesize this variance is due to many SAE features encoding the same underlying $U$ and variance in which of these features are selected. Comparing SAE versus SAE$_{select}$ across all CEM settings (\cref{tab:pipeline-comparison-20ng}), the latter has similar bias metrics but retains far more treated and control units after matching. With DoubleML, selection offers little benefit for estimating ATT, and we hypothesize that the same effect is achieved implicitly by learned feature weighting in the nuisance functions. We further investigate the selection step in \cref{subsec:selection-step}.

\begin{figure*}[ht]
    \centering
    \begin{subfigure}{0.48\textwidth}
        \centering
        \includegraphics[width=\textwidth]{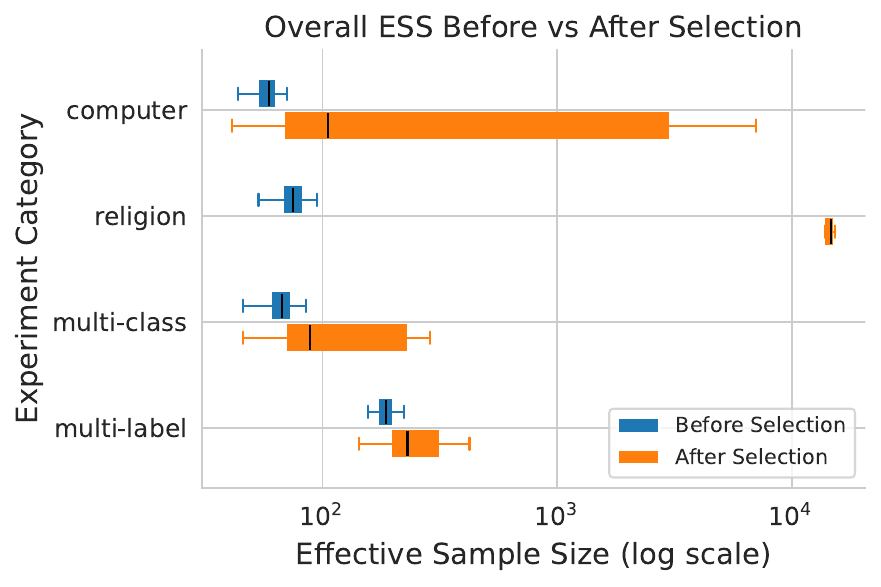}
        \caption{ESS distribution across simulations}
        \label{fig:ess_across_sims}
    \end{subfigure}
    \hfill
    \begin{subfigure}{0.48\textwidth}
        \centering
        \includegraphics[width=\textwidth]{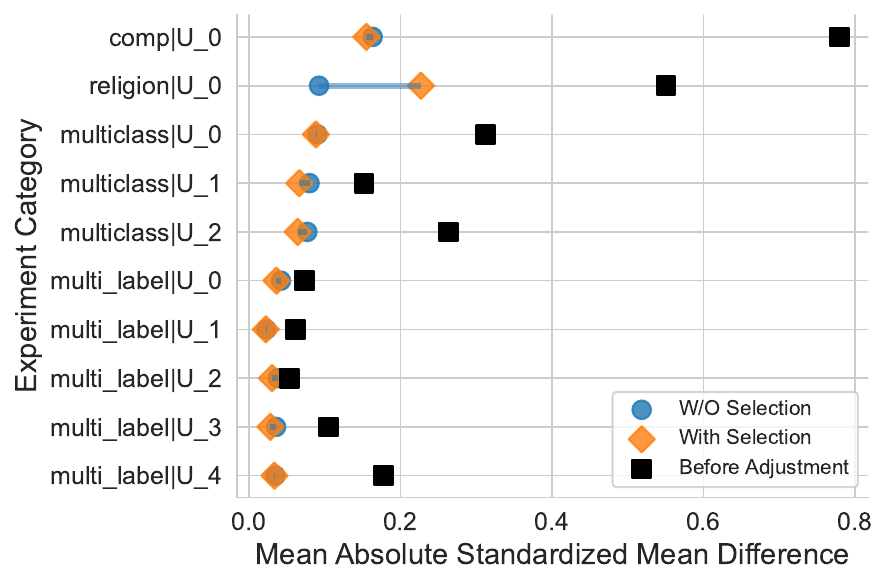}
        \caption{Average |SMD| on $U$}
        \label{fig:mean_smd}
    \end{subfigure}    

    \caption{Overlap support versus balance comparison, before and after feature selection on SAEs using the CEM estimator. For these diagnostics, we have access to the oracle confounder $U$.}
    \label{fig:overlap v.s. balance}
\end{figure*}

\subsection{Results for multi-label (EURLEX)}\label{subsec:eurlex}
Although the results for 20NG consistently showed the performance gains of SAEs, the results for the multi-label setting with EURLEX are more mixed~(\cref{tab:pipeline-eurlex}). We see that SAE+CEM (with and without selection) have low bias and high coverage but matching retains less than 1\% of treated units across both settings. In contrast, TIRM also achieves similar bias metrics, but has less coverage and retains far more treated units ($\approx21\%$). 

Similar to \cref{sec:results_20ng}, both STM and SAE when plugged into DoubleML reduce bias from the unadjusted estimate (0.2863) with biases of 0.1461 and 0.1324 respectively but the bias is still concernedly far from zero. When inspecting the DoubleML underlying nuisance estimators, we find the treatment classifier converges but there is a lack of convergence in the outcome estimator. 
Our hypothesis is this might be a finite data issue. In Appendix \cref{fig:eurlex-confounder-distribution} we show the distribution across the multi-label confounders. Although our dataset has strict overlap, many of the confounding buckets have ``lack of common support'', which \citet{hill2013assessing} define as neighborhoods of covariate space in which there are not sufficient numbers of treated and control units in the finite data sample. We find this an extremely interesting setting in which matching results in low bias but DoubleML results in high bias given the same input representations. We leave to future work deeper investigations of this phenomena and mitigation of lack of common support, e.g., building from \citet{hill2013assessing,oberst2020characterization}.

\subsection{Analysis of Selection Step}\label{subsec:selection-step}
We conduct additional analysis of the step in our pipeline that selects a subset of SAE features for overlap and balance diagnostics when using CEM estimator. First, in \cref{fig:overlap v.s. balance}, we report the effective sample size (ESS) across simulations before and after selection. For all confounding settings, selection increases ESS substantially, corroborating the results in Tab.~\cref{tab:pipeline-comparison-20ng,tab:pipeline-eurlex}, which show that selection retains a greater percentage of data and improves overlap in matching.
Second, we investigate if selection improves or degrades the balance of true confounders in the matched datasets. For balance checking, we employ standardized mean deviation where we compute the absolute $|SMD|$ on the true confounders, $U$ after adjusting using SAE and $SAE_{select}$. In \cref{fig:mean_smd}, we find selection step maintains a better balance across true confounding variables with the lower $|SMD|$ for all settings except $U=religion$. However, even for $U=religion$, the SMD after selection is still maintained in a low reasonable area around $0.2$ that is well below the unadjusted SMD (0.55). 

\begin{figure}[h]
    \centering

    \begin{subfigure}{\linewidth}
        \centering
        \includegraphics[width=\linewidth]{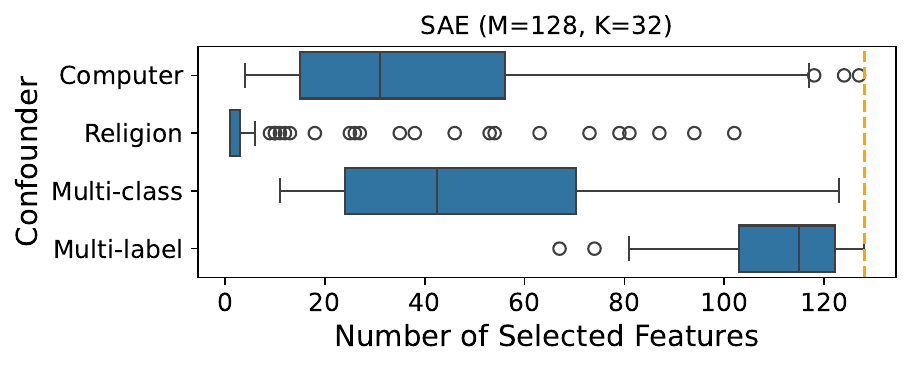}
        \caption{SAE ($M=128, K=32$) }
        \label{fig:n_selected_m128}
    \end{subfigure}

    \vspace{0.5em}

    \begin{subfigure}{\linewidth}
        \centering
        \includegraphics[width=\linewidth]{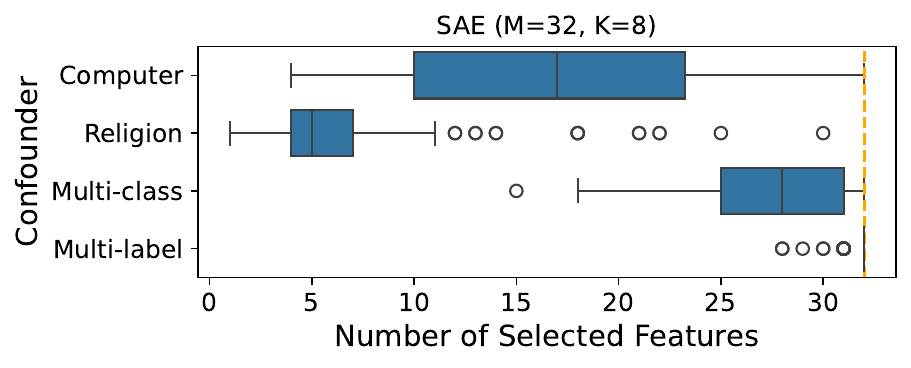}
        \caption{SAE ($M=32, K=8$) }
        \label{fig:n_selected_m32}
    \end{subfigure}

    \caption{Distribution of number of selected SAE features under different confounding settings across 100 simulations. 
    }
    \label{fig:n_selected_features}
\end{figure}

\subsection{Analysis of SAE Hyperparameters}\label{subec:hyperparams}

Next, we investigate the sensitivity of our SAE pipeline to hyperparameters, i.e., SAE $M$ and $K$ settings, and the opportunity for falsification of  the setting of M, as proposed in \S\ref{sec:methods_falsification}.
\cref{fig:n_selected_features} shows the distribution of the number of selected features using SAE $M=128$ and a smaller one $M=32$~(with a fixed $\frac{K}{M}$ ratio 0.25).
In $M=128$, more complex confounding settings like multi-class and multi-label in general select more features, where medians are above 40 out of 128. 
There is also variance in the number of selected features across simulations, which is consistent with the range of ESS in \cref{fig:ess_across_sims}. Between the binary confounders, the less frequent confounder ($U = \text{religion}$) has a larger variance in feature selection, suggesting SAE representations may encode these less-frequent features less atomically.

When we change the number of dimensions in the SAE representations to $M=32$, selection still reduces the feature set for binary confounders, but nearly all the neurons are selected in multi-class and multi-labeled settings. These results suggest $M=32$ is not sufficiently expressive to atomically capture these more complex label sets. They also demonstrate the simple empirical falsification check that selection offers: if nearly all of the neurons are picked, a larger SAE is likely needed.
  
\begin{figure}[h]
    \begin{subfigure}[c]{0.6\linewidth}
        \includegraphics[width=\linewidth]{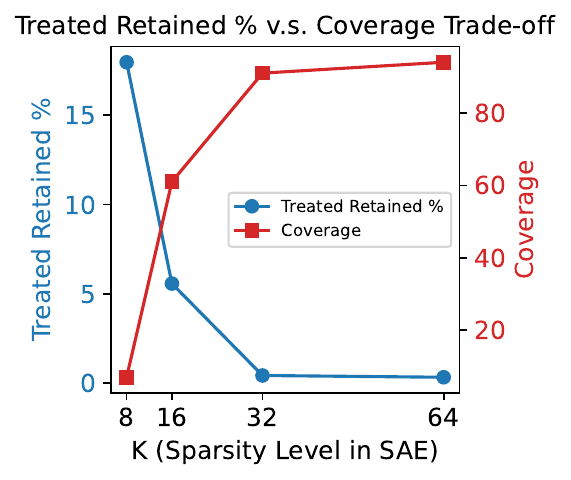}
    \caption{}
    \label{fig:M128_retain_coverage_tradeoff}
    \end{subfigure}
    \hfill
    \begin{subfigure}[c]{0.39\linewidth}
        \centering
        \resizebox{\linewidth}{!}{
        \begin{tabular}{lrr}
            \toprule
            K & Bias & RMSE \\
            \midrule
            8 & .036 & .038 \\
            16 &  .027 & .031 \\
            32 & .031 & .056\\
            64 & .028 & .125\\
            \bottomrule
        \end{tabular}}
        \caption{}
        \label{tab:multilabel-bias-rmse-diff-K}
    \end{subfigure}
    \caption{\textbf{SAE sparsity ($K$) ablation} in our multi-label setting. For, SAE$_{select}$ with dimension $M=128$, a smaller $K$ indicates greater sparsity of the model as only $K$ neurons are retained.
    }
    \label{fig:k_sparse_abalation}
\end{figure}

Furthermore, with an increasing $M$, the decision of how to set the sparsity $K$ becomes non-trivial. We conduct an ablation on different sparsity levels $K = (8, 16, 32, 64)$ for the multi-label setting and report results in \cref{fig:k_sparse_abalation}.
Greater sparsity generally improves RMSE and retains more data, but coverage improves with less sparsity, suggesting practitioners may need choose which metrics they most care about and set $K$ accordingly. Future work on hyperparameter setting would be beneficial for providing further guidance.

\subsection{Interpretability of SAE features}\label{sec:interpretability}
Finally, we qualitatively inspect SAE features to verify that they are interpretable and thus offer researchers greater oversight and opportunity for falsification than black-box methods, as proposed in \S\ref{sec:methods_falsification}.
Following~\citet{movva2025sparse}, for neurons identified by selection, we prompt an LLM with texts that have each neuron active, instructing it to generate a label for the selected features. We use the 128-feature SAE and look at neurons that are selected at least 40 times out of 100 simulations. In our data, the selected neurons are highly relevant to the true confounder we set, offering face validity. For instance, in a multi-class setting~(computer v.s. religion v.s. other), the top neurons include ``Contains explicity Bible scripture references'' and ``Mentions PC motherboard/bus hardware standards'', while the binary setting on religion only contains the former.
For the multi-label setting, all selected neurons appear in at least 40 simulations, so we manually check for neurons in $\geq 95$ simulations (25 in total) and find evidence for all five labels, for instance, ``Contains the exact title phrase `establishing the standard import values for determining the entry price of certain fruit and vegetables'' relates to Trade and Agri-foodstuffs. 
In a dataset where the selected neurons did not display content relevant to hypothesized confounders, this diagnostic could expose results as unreliable, falsifying the pipeline.

\section{Conclusion and Outlook}
In this work, we construct a novel pipeline and expand to complex causal simulation for adjusting for text as confounders by subsetting SAEs with statistically informed selection. Our exploration shows SAEs have the potential to aid in causal analyses and also opens channels for future work. SAEs may be sufficiently expressive to encode a wide range of confounding information beyond just topics, since the features they recover also include stylistic and semantic variation. Future work can investigate more diverse text confounders~(e.g., syntactic). Our results also suggest future investigations are needed to more deeply understand performance in multi-label settings. 
With respect to the adjustment pipeline, we empirically compare pipelines where the learned representations are plugged into downstream estimators. Future investigations of methods that jointly learn the representations and estimators \citep{veljanovski-wood-doughty-2024-doublelingo,veitch2020adapting} and systematic search on optimizing hyperparameters throughout the pipeline are needed.

\section*{Limitations}
Examining causal adjustment using semi-synthetic simulations involves multiple stages of parameter setting, such as, choices of confounder labels, different set-ups of confounding and treatment strength for the data-generating process, dimensionality of representations, etc., and our results may vary under other settings. To reduce this limitation, at each step, we carefully motivated decisions as well as followed best practices from prior work, and we report results across multiple simulation settings and datasets.

Moreover, to make comparisons among methods fair and principled, we used the same hyperparameter settings across pipelines. However, optimal parameter settings may vary across underlying representations. For instance, there may be different optimal binning rules of CEM for SAE representations and the constrained $(0,1)$ continuous-valued topic representation from STM. Nevertheless, SAEs seem to be more robust to such rules from our empirical results~(See \cref{app:results}).

Lastly, there are other SAE architectures that could be investigated. We defer these investigations to future work, as they can be easily integrated into our pipeline, and furthermore, comparisons of different SAEs offer minor refinements on our proposed work, rather than being necessary for our proposed adjustment approach to be usable.

\bibliography{custom, anthology-1}

\begin{thebibliography}{31}
\providecommand{\natexlab}[1]{#1}

\bibitem[{Bills et~al.(2023)Bills, Cammarata, Mossing, Tillman, Gao, Goh, Sutskever, Leike, Wu, and Saunders}]{Bills2023}
Steven Bills, Nick Cammarata, Dan Mossing, Henk Tillman, Leo Gao, Gabriel Goh, Ilya Sutskever, Jan Leike, Jeff Wu, and William Saunders. 2023.
\newblock \href {https://openai.com/index/language-models-can-explain-neurons-in-language-models/} {Language models can explain neurons in language models}.
\newblock OpenAI.

\bibitem[{Bricken et~al.(2023)Bricken, Templeton, Batson, Chen, Jermyn, Conerly, Turner, Anil, Denison, Askell, Lasenby, Wu, Kravec, Schiefer, Maxwell, Joseph, Tamkin, Nguyen, McLean, Burke, Hume, Carter, Henighan, and Olah}]{bricken2023towards}
Trenton Bricken, Adly Templeton, Joshua Batson, Brian Chen, Adam Jermyn, Tom Conerly, Nicholas~L Turner, Cem Anil, Carson Denison, Amanda Askell, Robert Lasenby, Yifan Wu, Shauna Kravec, Nicholas Schiefer, Tim Maxwell, Nicholas Joseph, Alex Tamkin, Karina Nguyen, Brayden McLean, and 5 others. 2023.
\newblock \href {https://transformer-circuits.pub/2023/monosemantic-features} {Towards monosemanticity: Decomposing language models with dictionary learning}.
\newblock Transformer Circuits Thread, Anthropic.
\newblock Accessed: 2026-05-20.

\bibitem[{Chalkidis et~al.(2021)Chalkidis, Fergadiotis, and Androutsopoulos}]{chalkidis-etal-2021-multieurlex}
Ilias Chalkidis, Manos Fergadiotis, and Ion Androutsopoulos. 2021.
\newblock \href {https://doi.org/10.18653/v1/2021.emnlp-main.559} {{M}ulti{EURLEX} - a multi-lingual and multi-label legal document classification dataset for zero-shot cross-lingual transfer}.
\newblock In \emph{Proceedings of the 2021 Conference on Empirical Methods in Natural Language Processing}, pages 6974--6996, Online and Punta Cana, Dominican Republic. Association for Computational Linguistics.

\bibitem[{Chen et~al.(2024)Chen, Bhattacharya, and Keith}]{chen2024proximal}
Jacob~M. Chen, Rohit Bhattacharya, and Katherine~A. Keith. 2024.
\newblock \href {https://openreview.net/forum?id=L4RwA0qyUd} {Proximal causal inference with text data}.
\newblock In \emph{The Thirty-eighth Annual Conference on Neural Information Processing Systems}.

\bibitem[{Chernozhukov et~al.(2017)Chernozhukov, Chetverikov, Demirer, Duflo, Hansen, and Newey}]{chernozhukov2017double}
Victor Chernozhukov, Denis Chetverikov, Mert Demirer, Esther Duflo, Christian Hansen, and Whitney Newey. 2017.
\newblock \href {https://doi.org/10.1257/aer.p20171038} {Double/debiased/neyman machine learning of treatment effects}.
\newblock \emph{American Economic Review}, 107(5):261–65.

\bibitem[{Choi et~al.(2025)Choi, Lim, Schneider, and Choo}]{Choi2025}
Jinho Choi, Hyesu Lim, Steffen Schneider, and Jaegul Choo. 2025.
\newblock \href {https://proceedings.neurips.cc/paper_files/paper/2025/file/0ddf14f20994636eeecc3d96fa8545cf-Paper-Conference.pdf} {Conceptscope: Characterizing dataset bias via disentangled visual concepts}.
\newblock In \emph{Advances in Neural Information Processing Systems}, volume~38, pages 9604--9639. Curran Associates, Inc.

\bibitem[{D’Amour et~al.(2021)D’Amour, Ding, Feller, Lei, and Sekhon}]{d2021overlap}
Alexander D’Amour, Peng Ding, Avi Feller, Lihua Lei, and Jasjeet Sekhon. 2021.
\newblock \href {https://doi.org/10.1016/j.jeconom.2019.10.014} {Overlap in observational studies with high-dimensional covariates}.
\newblock \emph{Journal of Econometrics}, 221(2):644--654.

\bibitem[{Feder et~al.(2022)Feder, Keith, Manzoor, Pryzant, Sridhar, Wood-Doughty, Eisenstein, Grimmer, Reichart, Roberts, Stewart, Veitch, and Yang}]{feder-etal-2022-causal}
Amir Feder, Katherine~A. Keith, Emaad Manzoor, Reid Pryzant, Dhanya Sridhar, Zach Wood-Doughty, Jacob Eisenstein, Justin Grimmer, Roi Reichart, Margaret~E. Roberts, Brandon~M. Stewart, Victor Veitch, and Diyi Yang. 2022.
\newblock \href {https://doi.org/10.1162/tacl_a_00511} {Causal inference in natural language processing: Estimation, prediction, interpretation and beyond}.
\newblock \emph{Transactions of the Association for Computational Linguistics}, 10:1138--1158.

\bibitem[{Gao et~al.(2025)Gao, la~Tour, Tillman, Goh, Troll, Radford, Sutskever, Leike, and Wu}]{gao2025scaling}
Leo Gao, Tom~Dupre la~Tour, Henk Tillman, Gabriel Goh, Rajan Troll, Alec Radford, Ilya Sutskever, Jan Leike, and Jeffrey Wu. 2025.
\newblock \href {https://openreview.net/forum?id=tcsZt9ZNKD} {Scaling and evaluating sparse autoencoders}.
\newblock In \emph{The Thirteenth International Conference on Learning Representations}.

\bibitem[{Hern{\'a}n(2016)}]{hernan2016does}
Miguel~A. Hern{\'a}n. 2016.
\newblock \href {https://doi.org/10.1016/j.annepidem.2016.08.016} {Does water kill? {A} call for less casual causal inferences}.
\newblock \emph{Annals of Epidemiology}, 26(10):674--680.

\bibitem[{Hill and Su(2013)}]{hill2013assessing}
Jennifer Hill and Yu-Sung Su. 2013.
\newblock \href {https://doi.org/10.1214/13-AOAS630} {{Assessing lack of common support in causal inference using Bayesian nonparametrics: Implications for evaluating the effect of breastfeeding on children’s cognitive outcomes}}.
\newblock \emph{The Annals of Applied Statistics}, 7(3):1386 -- 1420.

\bibitem[{Huben et~al.(2024)Huben, Cunningham, Smith, Ewart, and Sharkey}]{Huben2024}
Robert Huben, Hoagy Cunningham, Logan Smith, Aidan Ewart, and Lee Sharkey. 2024.
\newblock \href {https://proceedings.iclr.cc/paper_files/paper/2024/file/1fa1ab11f4bd5f94b2ec20e794dbfa3b-Paper-Conference.pdf} {Sparse autoencoders find highly interpretable features in language models}.
\newblock In \emph{International Conference on Learning Representations}, volume 2024, pages 7827--7845.

\bibitem[{Iacus et~al.(2012)Iacus, King, and Porro}]{iacus2012}
Stefano~M. Iacus, Gary King, and Giuseppe Porro. 2012.
\newblock \href {https://doi.org/10.1093/pan/mpr013} {Causal inference without balance checking: Coarsened exact matching}.
\newblock \emph{Political Analysis}, 20(1):1–24.

\bibitem[{Jiang et~al.(2025)Jiang, Sun, Smith, and Nanda}]{jiang2025towards}
Nicholas Jiang, Xiaoqing Sun, Lewis Smith, and Neel Nanda. 2025.
\newblock \href {https://openreview.net/forum?id=mqJbhBMFm5} {Towards data-centric interpretability with sparse autoencoders}.
\newblock In \emph{Mechanistic Interpretability Workshop at NeurIPS 2025}.

\bibitem[{Keith et~al.(2023)Keith, Feldman, Jurgens, Bragg, and Bhattacharya}]{keithrct}
Katherine~A. Keith, Sergey Feldman, David Jurgens, Jonathan Bragg, and Rohit Bhattacharya. 2023.
\newblock \href {https://openreview.net/forum?id=F74ZZk5hPa} {{RCT} rejection sampling for causal estimation evaluation}.
\newblock \emph{Transactions on Machine Learning Research}.

\bibitem[{Keith et~al.(2020)Keith, Jensen, and O{'}Connor}]{keith-etal-2020-text}
Katherine~A. Keith, David Jensen, and Brendan O{'}Connor. 2020.
\newblock \href {https://doi.org/10.18653/v1/2020.acl-main.474} {Text and causal inference: A review of using text to remove confounding from causal estimates}.
\newblock In \emph{Proceedings of the 58th Annual Meeting of the Association for Computational Linguistics}, pages 5332--5344, Online. Association for Computational Linguistics.

\bibitem[{Lang(1995)}]{Lang95}
Ken Lang. 1995.
\newblock Newsweeder: learning to filter netnews.
\newblock In \emph{Proceedings of the Twelfth International Conference on International Conference on Machine Learning}, ICML'95, page 331–339, San Francisco, CA, USA. Morgan Kaufmann Publishers Inc.

\bibitem[{Makhzani and Frey(2014)}]{makhzani2013k}
Alireza Makhzani and Brendan~J. Frey. 2014.
\newblock \href {http://arxiv.org/abs/1312.5663} {k-{S}parse autoencoders}.
\newblock In \emph{2nd International Conference on Learning Representations, {ICLR} 2014, Banff, AB, Canada, April 14-16, 2014, Conference Track Proceedings}.

\bibitem[{Movva et~al.(2025)Movva, Peng, Garg, Kleinberg, and Pierson}]{movva2025sparse}
Rajiv Movva, Kenny Peng, Nikhil Garg, Jon Kleinberg, and Emma Pierson. 2025.
\newblock \href {https://openreview.net/forum?id=4R0pugRyN5} {Sparse autoencoders for hypothesis generation}.
\newblock In \emph{Forty-second International Conference on Machine Learning}.

\bibitem[{Oberst et~al.(2020)Oberst, Johansson, Wei, Gao, Brat, Sontag, and Varshney}]{oberst2020characterization}
Michael Oberst, Fredrik Johansson, Dennis Wei, Tian Gao, Gabriel Brat, David Sontag, and Kush Varshney. 2020.
\newblock \href {https://proceedings.mlr.press/v108/oberst20a.html} {Characterization of overlap in observational studies}.
\newblock In \emph{Proceedings of the Twenty Third International Conference on Artificial Intelligence and Statistics}, volume 108 of \emph{Proceedings of Machine Learning Research}, pages 788--798. PMLR.

\bibitem[{Peng et~al.(2026)Peng, Movva, Kleinberg, Pierson, and Garg}]{peng2025use}
Kenny Peng, Rajiv Movva, Jon Kleinberg, Emma Pierson, and Nikhil Garg. 2026.
\newblock \href {https://openreview.net/forum?id=x1Px74tbvs} {Position: Use sparse autoencoders to discover unknowns}.

\bibitem[{Roberts et~al.(2020)Roberts, Stewart, and Nielsen}]{roberts2020adjusting}
Margaret~E. Roberts, Brandon~M. Stewart, and Richard~A. Nielsen. 2020.
\newblock \href {https://doi.org/10.1111/ajps.12526} {Adjusting for confounding with text matching}.
\newblock \emph{American Journal of Political Science}, 64(4):887--903.

\bibitem[{Roberts et~al.(2013)Roberts, Stewart, Tingley, Airoldi et~al.}]{roberts2013structural}
Margaret~E Roberts, Brandon~M Stewart, Dustin Tingley, Edoardo~M Airoldi, and 1 others. 2013.
\newblock \href {https://bstewart.scholar.princeton.edu/sites/g/files/toruqf4016/files/bstewart/files/stmnips2013.pdf} {The structural topic model and applied social science}.
\newblock In \emph{Advances in neural information processing systems workshop on topic models: computation, application, and evaluation}, volume~4, pages 1--20. Harrahs and Harveys, Lake Tahoe.

\bibitem[{Schulte et~al.(2025)Schulte, R{\"u}gamer, and Nagler}]{schulte2025adjustment}
Rickmer Schulte, David R{\"u}gamer, and Thomas Nagler. 2025.
\newblock \href {https://openreview.net/forum?id=s2HTzQ0sMY} {Adjustment for confounding using pre-trained representations}.
\newblock In \emph{ICLR 2025 Workshop on Foundation Models in the Wild}.

\bibitem[{Stuart(2010)}]{Stuart_2010}
Elizabeth~A. Stuart. 2010.
\newblock \href {https://doi.org/10.1214/09-sts313} {Matching methods for causal inference: A review and a look forward}.
\newblock \emph{Statistical Science}, 25(1).

\bibitem[{Tamarchenko(2023)}]{tamarchenko2023combining}
Elijah Tamarchenko. 2023.
\newblock Combining optimal adjustment set selection and post selection inference in unknown causal graphs.
\newblock Bachelor's thesis, Williams College, Williamstown, MA, May.
\newblock Advisor: Rohit Bhattacharya.

\bibitem[{Templeton et~al.(2024)Templeton, Conerly, Marcus, Lindsey, Bricken, Chen et~al.}]{templeton2024scaling}
A~Templeton, T~Conerly, J~Marcus, J~Lindsey, T~Bricken, B~Chen, and 1 others. 2024.
\newblock \href {https://transformer-circuits.pub/2024/scaling-monosemanticity/,} {Scaling monosemanticity: Extracting interpretable features from claude 3 sonnet. transformer circuits thread}.
\newblock Transformer Circuits Thread, Anthropic.

\bibitem[{Veitch et~al.(2020)Veitch, Sridhar, and Blei}]{veitch2020adapting}
Victor Veitch, Dhanya Sridhar, and David Blei. 2020.
\newblock \href {https://proceedings.mlr.press/v124/veitch20a.html} {Adapting text embeddings for causal inference}.
\newblock In \emph{Proceedings of the 36th Conference on Uncertainty in Artificial Intelligence (UAI)}, volume 124 of \emph{Proceedings of Machine Learning Research}, pages 919--928. PMLR.

\bibitem[{Veljanovski and Wood-Doughty(2024)}]{veljanovski-wood-doughty-2024-doublelingo}
Marko Veljanovski and Zach Wood-Doughty. 2024.
\newblock \href {https://doi.org/10.18653/v1/2024.naacl-short.71} {{D}ouble{L}ingo: Causal estimation with large language models}.
\newblock In \emph{Proceedings of the 2024 Conference of the North American Chapter of the Association for Computational Linguistics: Human Language Technologies (Volume 2: Short Papers)}, pages 799--807, Mexico City, Mexico. Association for Computational Linguistics.

\bibitem[{Wood-Doughty et~al.(2018)Wood-Doughty, Shpitser, and Dredze}]{wood-doughty-etal-2018-challenges}
Zach Wood-Doughty, Ilya Shpitser, and Mark Dredze. 2018.
\newblock \href {https://doi.org/10.18653/v1/D18-1488} {Challenges of using text classifiers for causal inference}.
\newblock In \emph{Proceedings of the 2018 Conference on Empirical Methods in Natural Language Processing}, pages 4586--4598, Brussels, Belgium. Association for Computational Linguistics.

\bibitem[{Zhang et~al.(2023)Zhang, Kennard, Smith, McFarland, McCallum, and Keith}]{zhang-etal-2023-causal-matching}
Raymond Zhang, Neha~Nayak Kennard, Daniel Smith, Daniel McFarland, Andrew McCallum, and Katherine~A. Keith. 2023.
\newblock \href {https://doi.org/10.18653/v1/2023.findings-acl.83} {Causal matching with text embeddings: A case study in estimating the causal effects of peer review policies}.
\newblock In \emph{Findings of the Association for Computational Linguistics: ACL 2023}, pages 1284--1297, Toronto, Canada. Association for Computational Linguistics.

\end{thebibliography}

\appendix
\section{Causal identification assumptions}\label{app:identification-assumptions}

From \cref{identification-assumptions}, we listed the four causal identification assumptions that cannot be verified from data alone:
\begin{enumerate}

    \item \emph{No unmeasured confounding.} The unobserved variables $U$ constitute a valid backdoor adjustment set for the effect of $T$ on $Y$.
    \item \emph{Overlap.} Every unit in the population has a nonzero probability of receiving each treatment level given its covariates, $0 < Pr(T=t \mid U) < 1, \forall t \in T$
    \item \emph{Consistency.} The outcome observed for each unit at treatment level $ t \in \{0, 1\}$ is identical to the outcome we would have observed had the unit been assigned to treatment level $t$.\footnote{Although consistency may sound like a trivial assumption to many readers unfamiliar with causal inference, we recommend the reader to \citet{hernan2016does}.}
    \item \emph{Homogeneous effects.} The treatment effect is constant across units; we leave heterogeneous effects to future work. 
\end{enumerate}

\section{Data Statistics}\label{app:data}
In this section, we describe more statistics and pre-processing on the two datasets we use for semi-synthetic evaluations.

\paragraph{20NG}
The dataset originally contains $20$ labels that are merged into $10$ labels used in our experiments as follows,

\begin{enumerate}
    \item computer: comp.graphics, comp.os.ms-windows.misc, comp.sys.ibm.pc.hardware, comp.sys.mac.hardware, comp.windows.x
    \item politics: talk.politics.guns, talk.politics.misc, talk.politics.mideast
    \item religion: alt.atheism, soc.religion.christian, talk.religion.misc
    \item sport: rec.sport.baseball, rec.sport.hockey
    \item automobile: rec.autos
    \item cryptography:sci.crypt
    \item medicine: sci.med
    \item forsale: misc.forsale
    \item electronics: sci.electronics
    \item space: sci.space
\end{enumerate}

\begin{figure}[h]
    \centering
    \includegraphics[width=\linewidth]{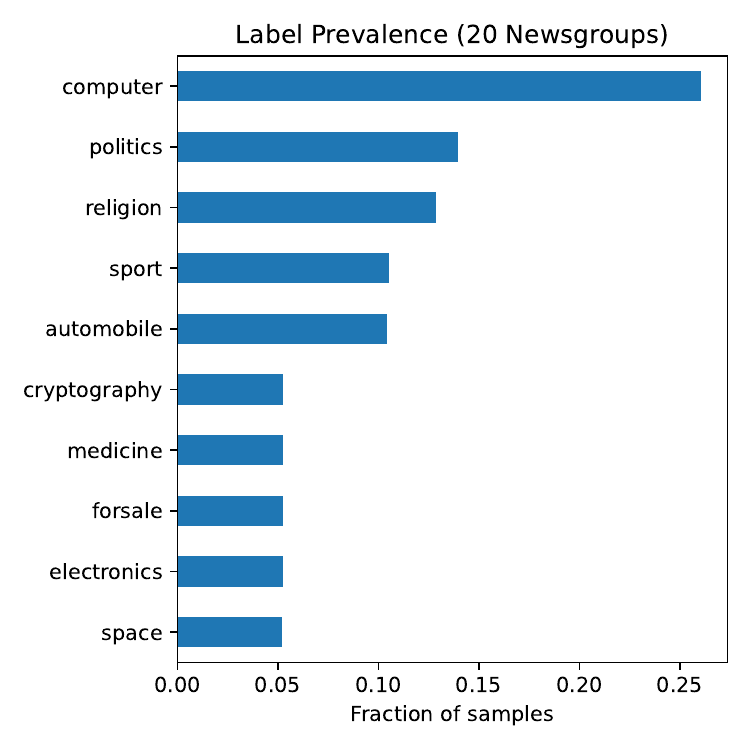}
    \caption{Label prevalence in 20NG data by fraction of total samples.}
    \label{fig:20news-prevalence}
\end{figure}

We present the prevalence of data labels of 20NG data by fraction in \cref{fig:20news-prevalence}.

\begin{figure}[h]
    \centering
    \includegraphics[width=\linewidth]{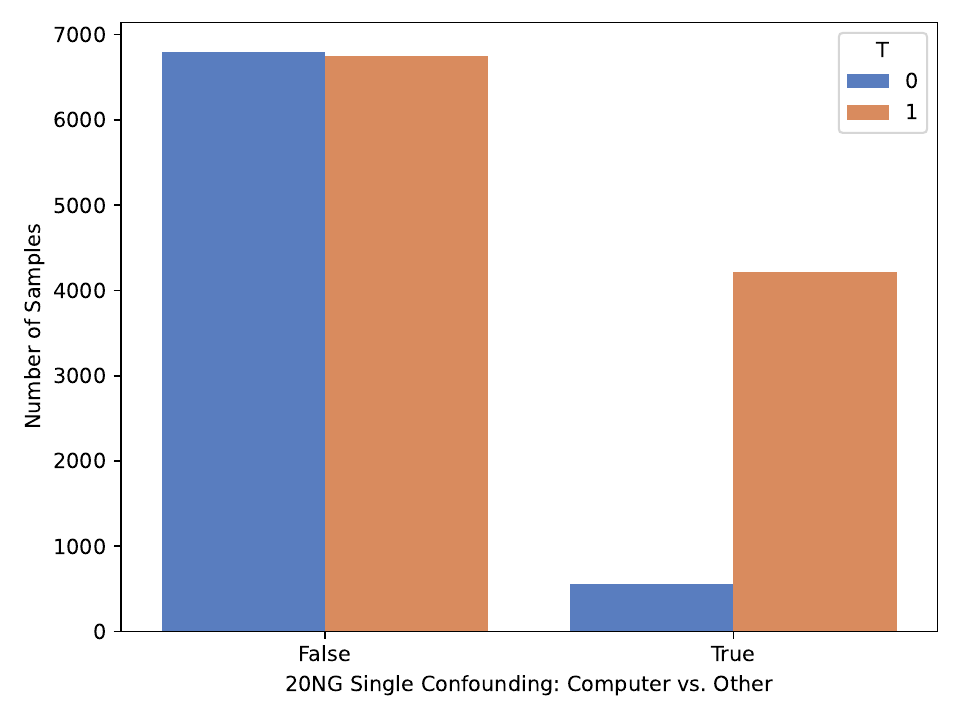}
    \caption{Treatment Assignment on 20NG computer}
    \label{fig:20news_comp_simulation}
\end{figure}

\begin{figure}[h]
    \centering
    \includegraphics[width=\linewidth]{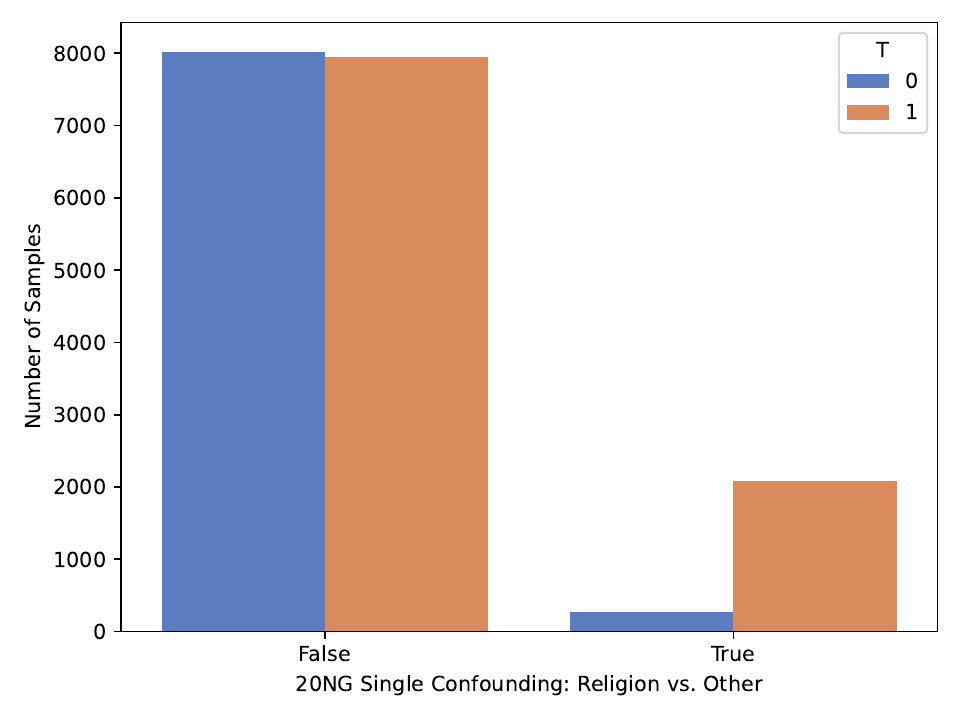}
    \caption{Treatment Assignment on 20NG religion}
    \label{fig:20news_religion_simulation}
\end{figure}

\begin{figure}[htbp]
    \centering
    \includegraphics[width=\linewidth]{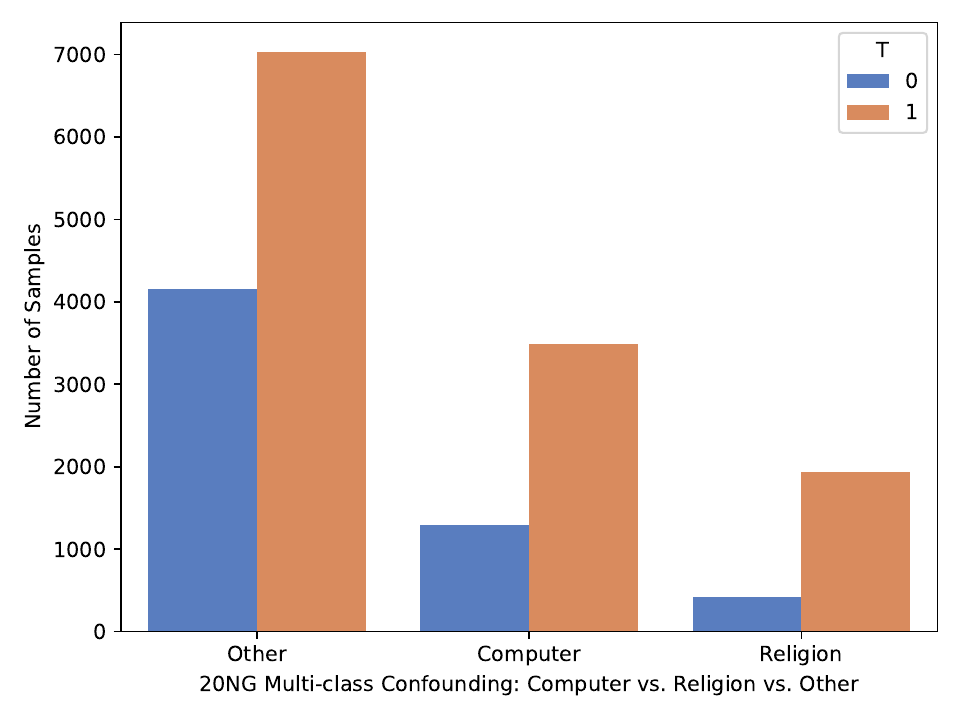}
    \caption{Treatment Assignment on 20NG multi-class (Computer, Religion, and Other)}
    \label{fig:20news_multiclass_simulation}
\end{figure}

In addition, as defined in our DGP~(\cref{subsec:dgp}), we illustrate an example treatment assignment distribution conditioned on the true confounder $U$ from our simulation for each confounding scenario which we can visualize the overlap. We plot 20NG computer in \cref{fig:20news_comp_simulation}, 20NG religion in \cref{fig:20news_religion_simulation}, and multi-class (computer, religion, and other) in \cref{fig:20news_multiclass_simulation}.

\begin{figure}[htbp]
    \centering
    \includegraphics[width=\linewidth]{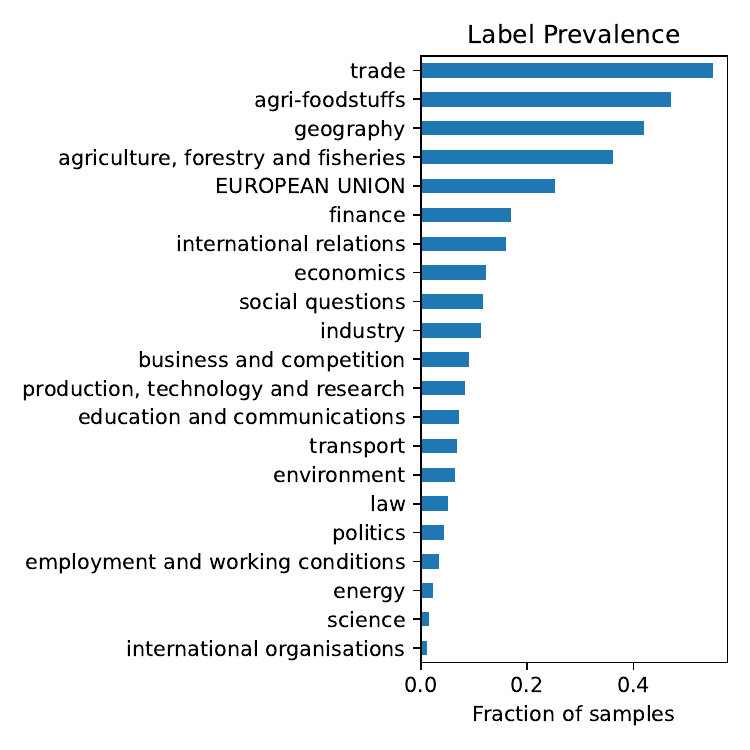}
    \caption{Label prevalence in original multi-labeled EURLEX data by fraction of total samples. We select the five most common labels to create our multi-labeled confounding experiment: ``trade'', ``agri-foodstuffs'', ``geography'', ``agriculture, forestry and fisheries'', and ``EUROPEAN UNION''.}
    \label{fig:eurlex-prevalence}
\end{figure}

\begin{figure*}[htbp]
    \centering
    \includegraphics[width=\linewidth]{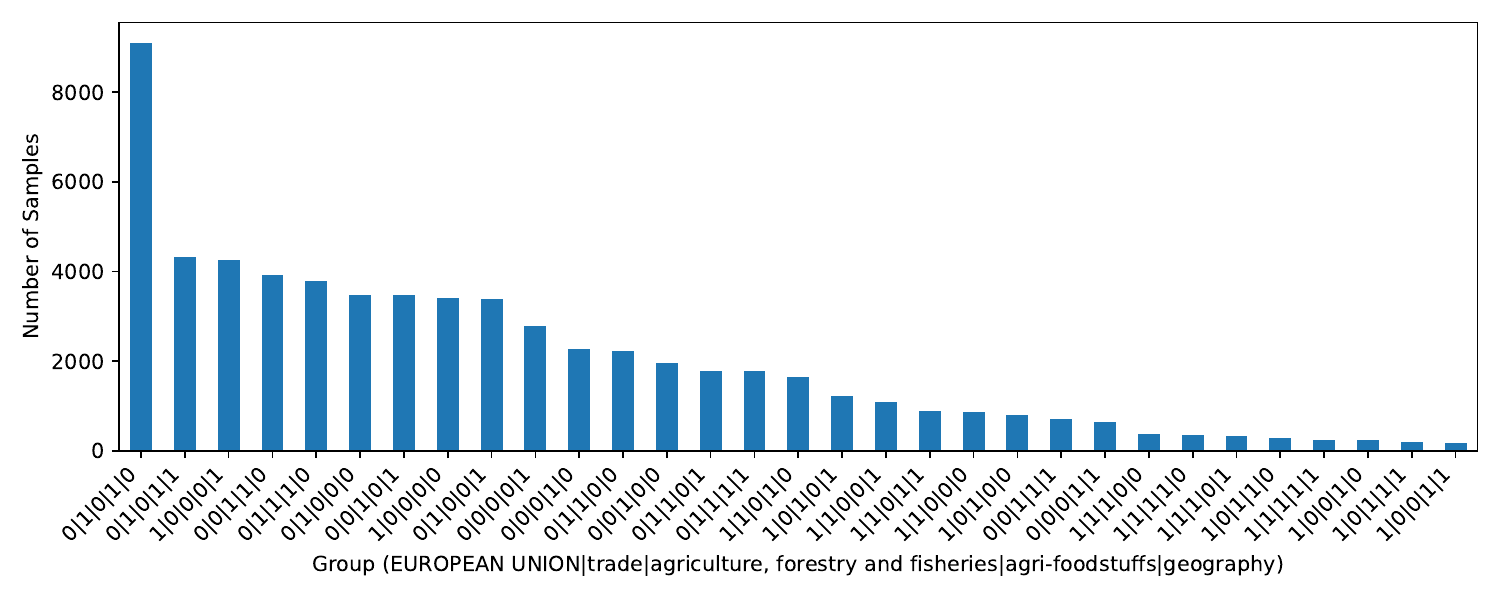}
    \caption{Distribution on multi-labeled true confounders in EURLEX}
    \label{fig:eurlex-confounder-distribution}
\end{figure*}

\paragraph{EURLEX} The original EURLEX dataset has 3-level hierarchical labels from coarse-grained to fine-grained. We use the coarsest level that contains 21 labels shown in \cref{fig:eurlex-prevalence} and select the top 5 labels: ``trade'', ``agri-foodstuffs'', ``geography'', agriculture, forestry and fisheries'', and ``EUROPEAN UNION''. In our multi-labeled experiment, these chosen labels are represented in a vector format. Finally, the distribution of the multi-labeled confounder of EURLEX in shown in \cref{fig:eurlex-confounder-distribution}.

\begin{figure*}
     \centering
     \includegraphics[width=\linewidth]{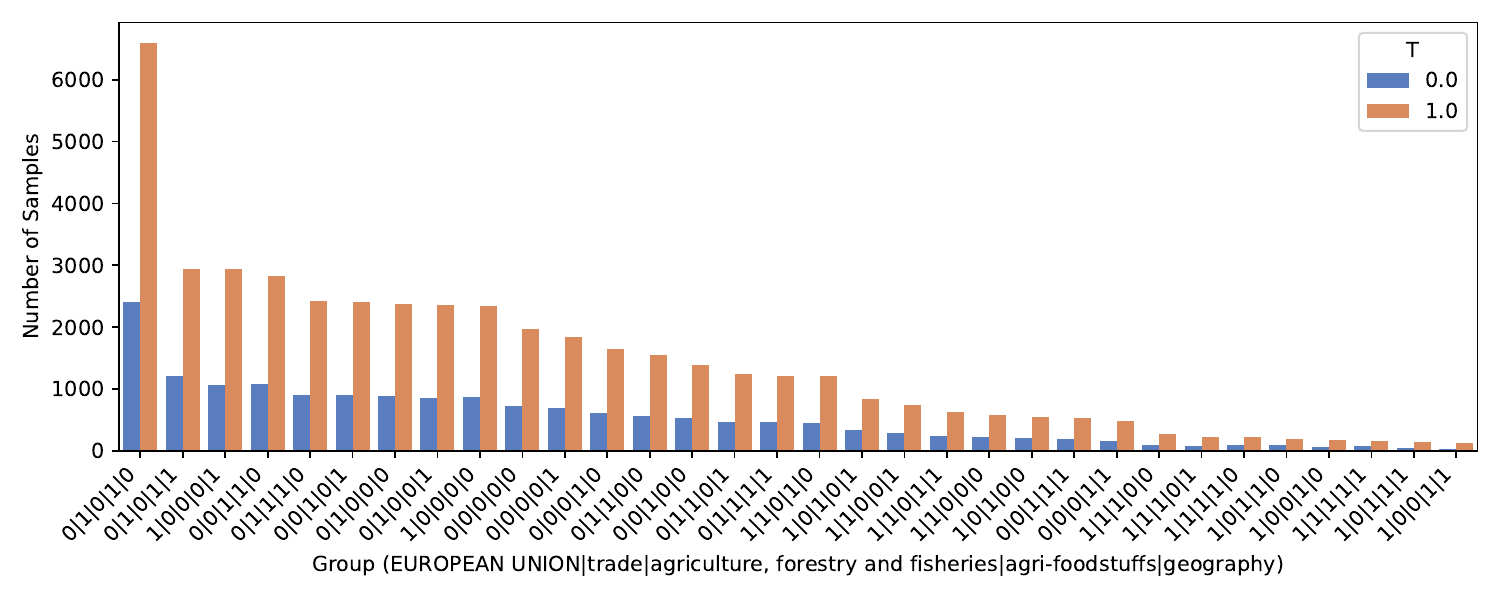}
     \caption{Treatment assignment on EURLEX}
     \label{fig:eurlex-treatment}
 \end{figure*}

 Furthermore, using one simulation from 100 simulations, the treatment distribution of multi-label EURLEX data is illustrated in \cref{fig:eurlex-treatment}. We can see that overlap satisfies but certain regions have poor overlap.

\section{Metrics Details}\label{sec:appendix:overlap}

Here we provide calculation details for the metrics. 
\paragraph{ATT}The Average Treatment Effect on the Treated (ATT) is the average effect of the treatment among treated units:
\[
\mathrm{ATT}
=
\mathbb{E}\left[ Y(1) - Y(0) \mid T = 1 \right],
\]
where $Y(1)$ and $Y(0)$ are the potential outcomes under treatment and control, and $T=1$ denotes treated units.

Assume we run $R$ simulations, with true ATT $\tau$, we compute the following empirical estimates for our method and baselines,
\begin{itemize}
    \item $\text{bias} = \frac{1}{R}\sum(\hat{\tau} - \tau)$
    \item $\text{RMSE} = \sqrt{\frac{1}{R}\sum(\hat{\tau} - \tau)^2}$
    \item coverage $= \frac{1}{R} \sum \tau \in CI$, the percentage of CI's that cover the true treatment effect.
\end{itemize}

\paragraph{Effective Sample Size~(ESS)} To assess covariate overlap between treated and control units, we report the effective sample size (ESS) implied by the balancing weights.

\[ESS = \dfrac{(\sum_i w_i)^2}{\sum_i w_i^2}\]
where $w_i$ is the weight for each observational unit. Next, we describe how the weights are computed in coarsened exact matching~(CEM)

\paragraph{Weights from CEM}
Recall that we have access to texts $D =\{d_1, \dots, d_N\}$ containing unobserved confounder(s) $U$ such as topics. For CEM, we bin the values of a covariate into several bins and perform exact match on these bins such that a bin containing only treatment units is dropped. Then, for an observational unit $d_i$, the weight is $0$ if it is not matched, otherwise, denote the bin of the unit as $b$ and the total number of units in this bin as $N_b$, the weight is calculated as follows,

\[w_i = \begin{cases}
    1, & T_i = 1 \\
    \frac{N-\sum_i^N T_i}{\sum_i^N T_i}\cdot \frac{N_b-\sum_i^{N_b} T_i}{\sum_i^{N_b} T_i}, & T_i = 0
\end{cases}\]
where the weight of a control unit ($T_i=0$) is the ratio between control and treatment units of the original population multiplied by the same ratio within the bin.

\paragraph{Standardized Mean Difference~(SMD)} We calculate the SMD of a true confounder $U$ as follows,
\[|SMD(U)| = |\dfrac{\bar{U}_{T=1} - \bar{U}_{T=0}}{sd_{pooled}}|\]

\[sd_{\mathrm{pooled}}
=
\sqrt{
\frac{\mathrm{Var}(U_{T=1})+\mathrm{Var}(U_{T=0})}{2}
}\]
Specifically, the means and variances are weighted such that the weights are all set to $1$ before matching (i.e. each observational unit has equal weight), and are computed as above after matching.

\section{Experiment Implementation}\label{app:experiment}
For the text embeddings used as input for SAE and Embed+DML, we use \texttt{openai-text-embedding-3-small}. Our selection step uses a 7:3 train-test split ratio and the train-test split remains the same across simulations. CEM's binning rule uses cut-points at $0$ and $50\%$ quantile of the positive values. 
For baseline TIRM, we run experiments in R with \texttt{stm} and \texttt{cem} libraries. 
For the rest baselines, we run experiments in python. We use package \texttt{doubleml} for double machine learning estimator implementation and \texttt{pymatchit} for Coarsened Exact Matching. We follow \texttt{hypothesaes} to implement SAE models.

We want to make a note that, because SAE does not require treatment and outcome variables, we train an SAE representation once and use it across all simulations. On the other hand, STMs in TIRM do depend on the treatment variable, requiring us to retrain the model for every simulation, making this approach more computationally expensive for semi-synthetic evaluations.

\section{Results: Complementary Figures and Tables}\label{app:results}
\paragraph{Unadjusted estimate} The unadjusted treatment effect is $\mathbb{E}[Y|T=1] - \mathbb{E}[Y|T=0]$ and in practice we calculate the difference of the mean of outcome variable between treatment and control samples.

\paragraph{Complementary results for main results tables} Specific to matching estimators, we report the percentage of retained treated samples~(T-ret.\%) in \cref{tab:pipeline-comparison-20ng} and \cref{tab:pipeline-eurlex} that indicates the proportion of treated samples retained after matching. Similarly we provide a complementary table about the control retained rate in \cref{tab:contro-retain}. They both indicate how much of the original treated(/control) population is actually supported by comparable controls(/treated) after matching.
\begin{table}[]
    \centering
    \resizebox{\linewidth}{!}{
    \begin{tabular}{lrrrr}
    \toprule
        & computer & religion & multi-class & multi-label \\
        \midrule
        TIRM &  2.59 & 2.72 & 2.80 & 25.18\\
        SAE & 0.42 & 0.48 & 0.59 & 0.58\\
        SAE$_{select}$ & 18.59 & 82.04 & 6.68 & 0.92\\
        \bottomrule
    \end{tabular}
    }
    \caption{Percentage of control retained for different confounding settings and pipeline using CEM estimator.}
    \label{tab:contro-retain}
\end{table}

In addition, in \cref{tab:pipeline-comparison-20ng} and \cref{tab:pipeline-eurlex}, we compare methods where the matching estimator uses a binning rule with $50\%$ quantile cutpoint on positive values. We also experiment with a cutpoint with $80\%$ quantile and present the results in \cref{tab:pipeline-comparison-m128-cutoff0.8}.

\begin{figure*}[htbp]
    \centering

    \begin{subfigure}[t]{0.98\linewidth}
        \centering
        \includegraphics[width=\linewidth]{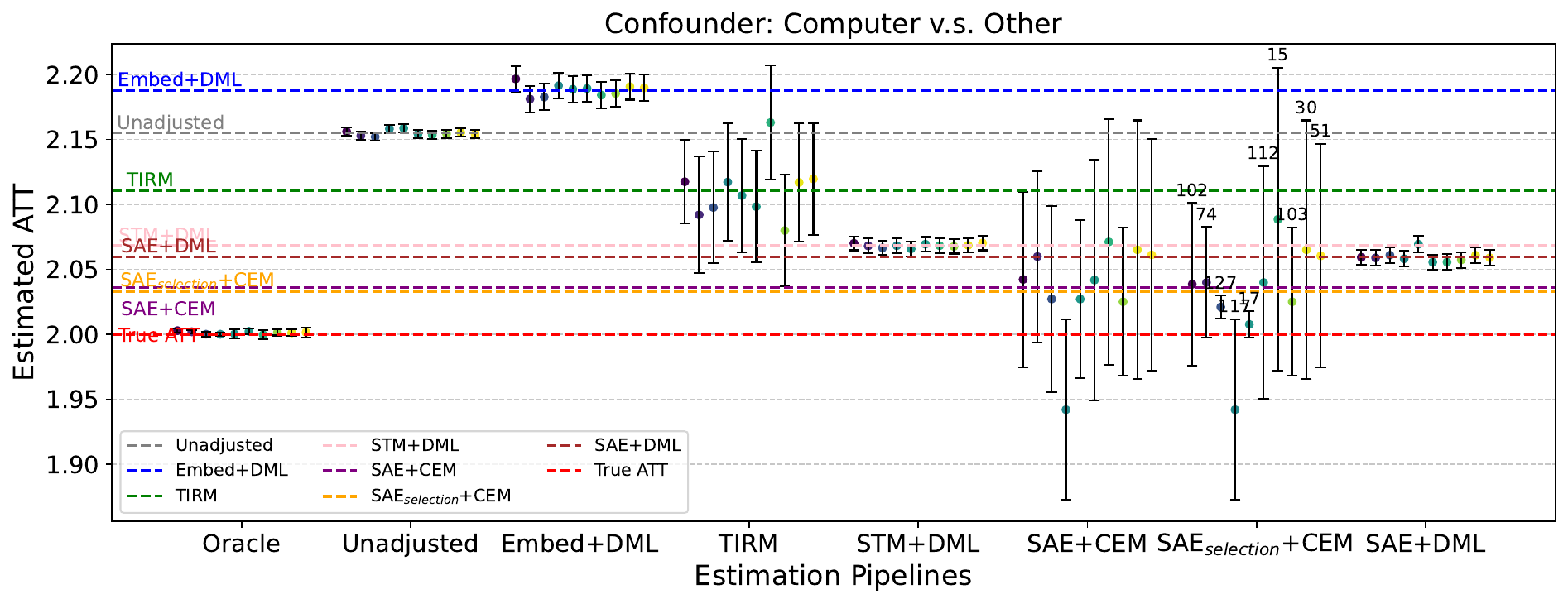}
        \caption{\textbf{Binary (computer)}. Simulations across 10 randomly selected seeds for the oracle binary confounder $U$ of computer versus other using the 20NewsGroups data.}
        \label{fig:comp_m128_comparison}
    \end{subfigure}

    \vspace{0.8em}

    \begin{subfigure}[t]{0.98\linewidth}
        \centering
        \includegraphics[width=\linewidth]{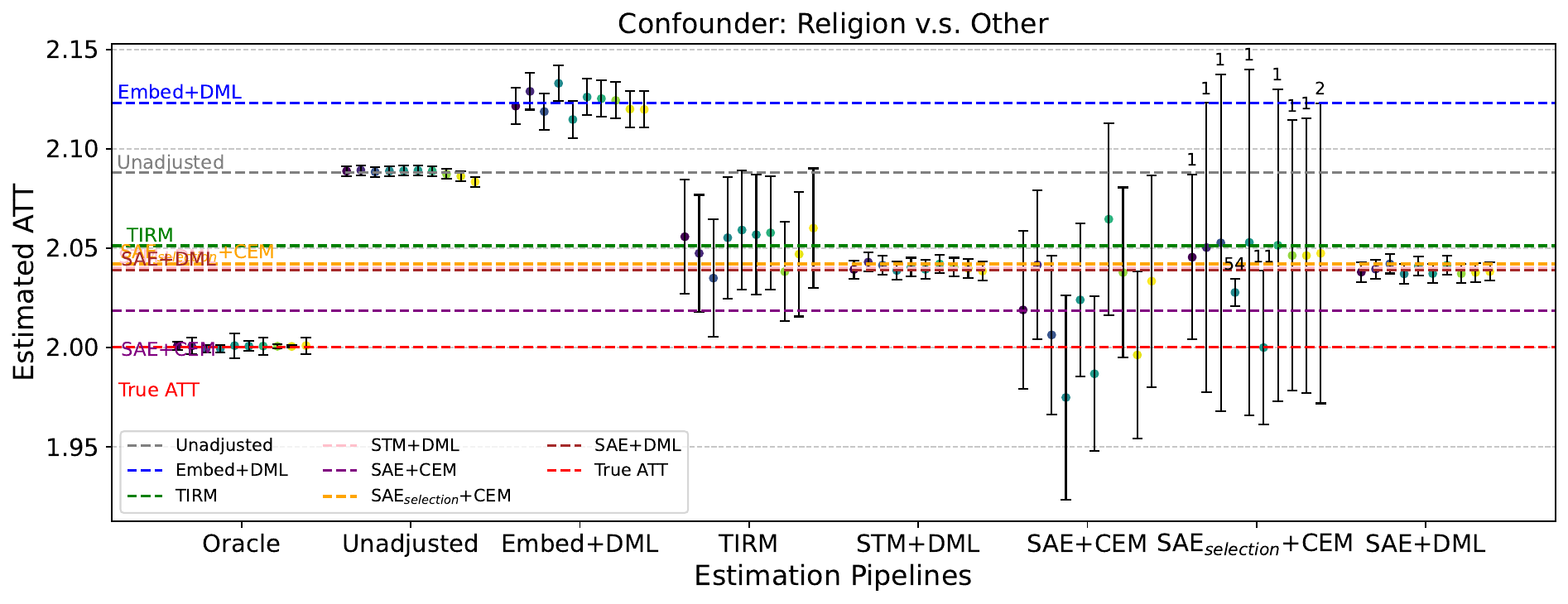}
        \caption{\textbf{Binary (religion)}. Simulations across 10 randomly selected seeds for the oracle binary confounder $U$ of religion versus other using the 20NewsGroups data.}
        \label{fig:religion_m128_comparison}
    \end{subfigure}

    \vspace{0.8em}

    \begin{subfigure}[t]{0.98\linewidth}
        \centering
        \includegraphics[width=\linewidth]{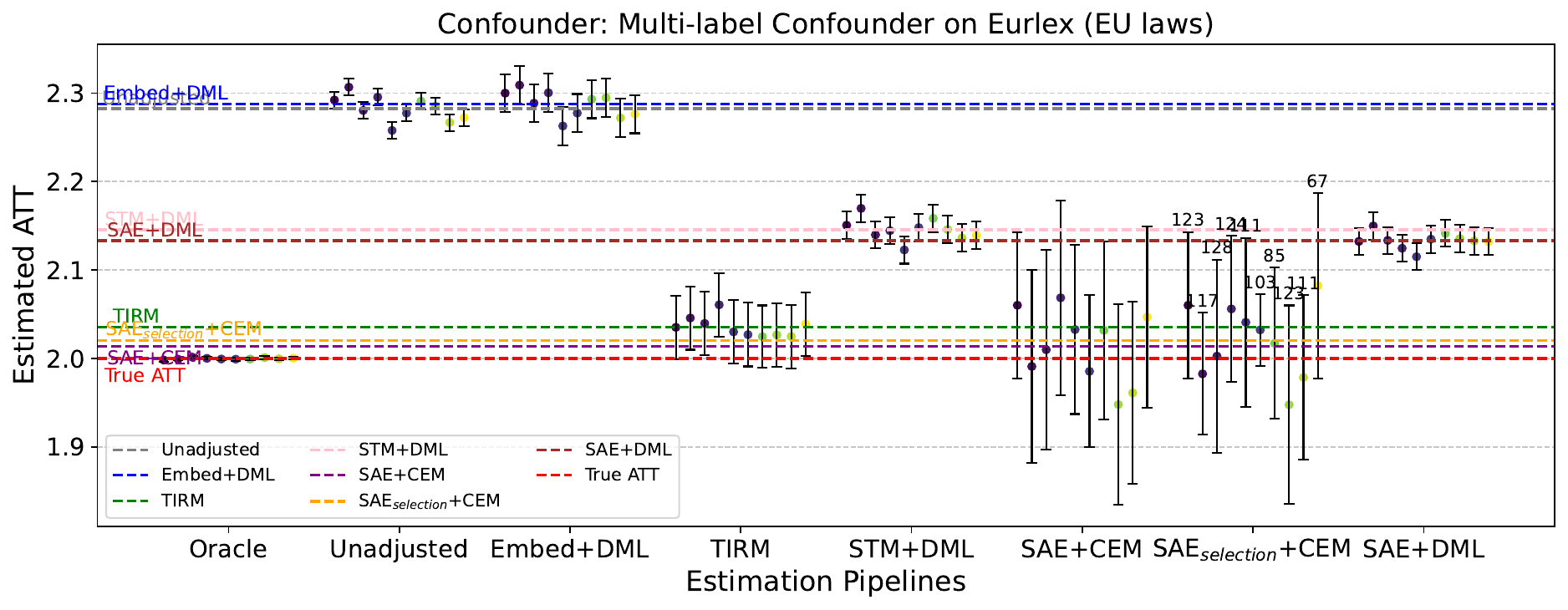}
        \caption{\textbf{EURLEX multi-label}. Simulations across 10 randomly selected seeds for the oracle multi-label confounder $U$ using the EURLEX data.}
        \label{fig:multilabel_m128_comparison}
    \end{subfigure}

    \caption{
    Simulation results across 10 out of 100 randomly selected seeds. Horizontal lines indicate the average estimate across these 10 simulations. DML denotes the DoubleML estimator. In SAE$_{selection}$+CEM, the number of selected features is listed on top of the confidence-interval bars.
    }
    \label{fig:m128_comparison_all}
\end{figure*}

\paragraph{Complementary figures for binary and multi-labeled confounding} Similar to \cref{fig:multiclass-m128}, we present the random 10-simulation adjustment performance across different pipelines figures for the other three confounding settings. We plot for binary~(computer) confounder in \cref{fig:comp_m128_comparison}, binary~(religion) confounder in \cref{fig:religion_m128_comparison}, and multi-labeled confounder in \cref{fig:multilabel_m128_comparison}.

\section{Interpretability of SAE features: Complementary Figures and Tables}\label{app:interpretability}
\begin{figure}[htbp]
    \centering
    \includegraphics[width=\linewidth]{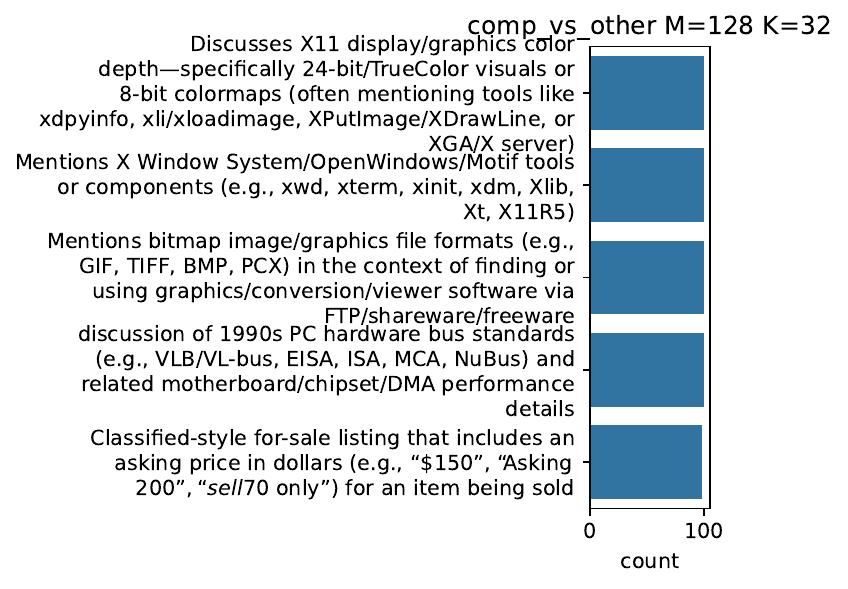}
    \caption{Top-5 most common neurons across 100 simulations for 20 news (computer) confounding}
    \label{fig:20ng-computer-top5neurons}
\end{figure}

\begin{figure}[htbp]
    \centering
    \includegraphics[width=\linewidth]{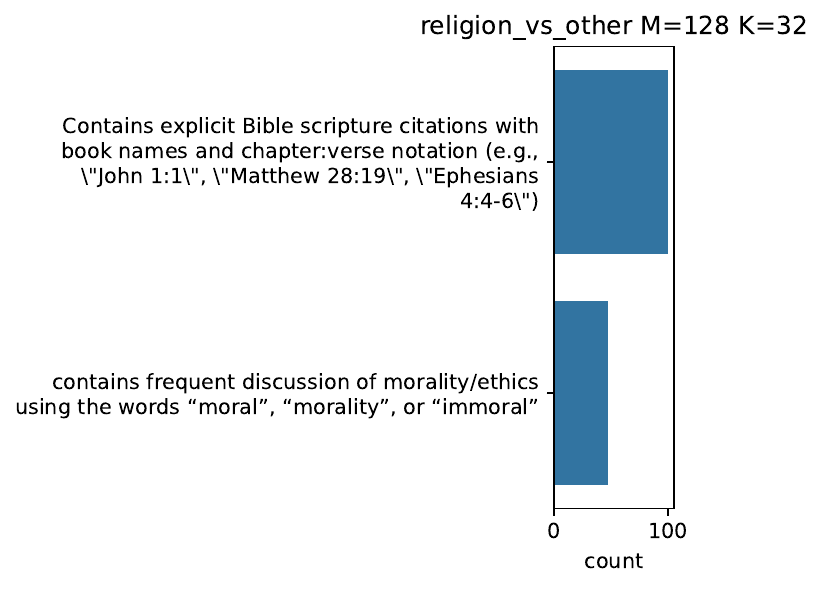}
    \caption{Top-5 most common neurons across 100 simulations for 20 news (religion) confounding}
    \label{fig:interpret-20news-religion}
\end{figure}

\begin{figure}[htbp]
    \centering
    \includegraphics[width=\linewidth]{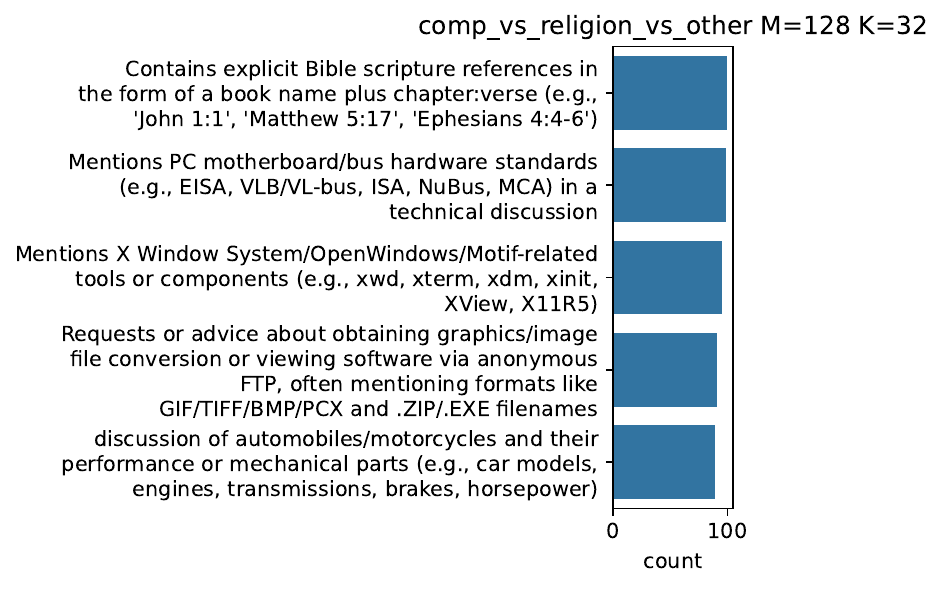}
    \caption{Top-5 most common neurons across 100 simulations for 20 news (computer and religion) multi-class confounding}
    \label{fig:interpret-20news-multiclass}
\end{figure}

\begin{figure}
    \centering
    \includegraphics[width=\linewidth]{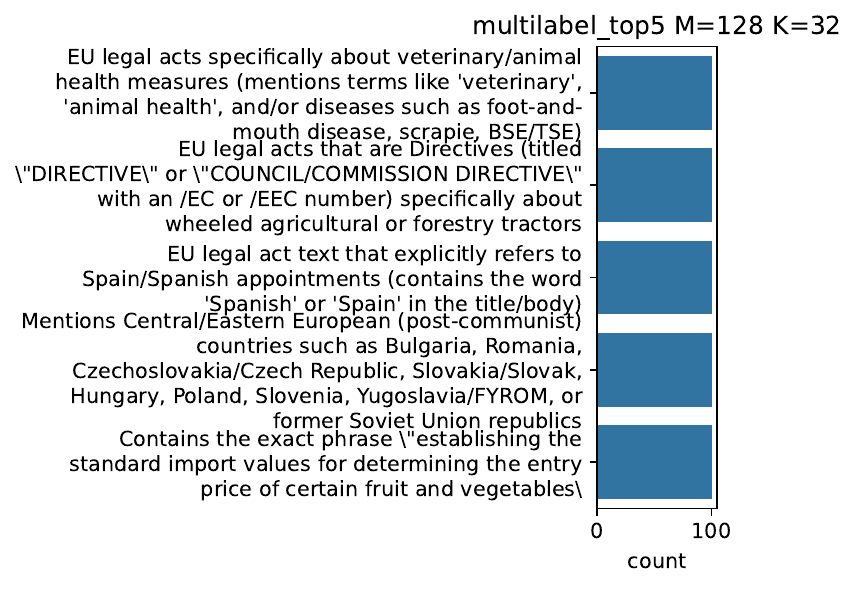}
    \caption{Top-5 most common neurons across 100 simulations for EURLEX multi-label confounding}
    \label{fig:interpret-eurlex-multilabel}
\end{figure}

We provide complementary figures and tables for the qualitative analysis in \cref{sec:interpretability} on SAE features for different confounding settings.
The prompting model is \texttt{gpt-5.2}.
For readability, we only plot the top five neurons that appear most frequent across 100 simulations in \cref{fig:20ng-computer-top5neurons} for 20NG computer, \cref{fig:interpret-20news-religion} for 20NG religion, \cref{fig:interpret-20news-multiclass} for 20NG multi-class, and \cref{fig:interpret-eurlex-multilabel} for EURLEX multi-label. A more comprehensive table for each confounding setting is provided.

\begin{table*}[ht]
\centering

\resizebox{\linewidth}{!}{
\begin{tabular}{llrrrrr}
\toprule
   Confounder & pipeline & Bias~$\downarrow$ & RMSE~$\downarrow$  & Coverage $\%$ ~$\uparrow$ & treated retained $\% \uparrow$ & control retained $\% \uparrow$\\
  \midrule
  
 Binary & TIRM & 0.0411 & 0.0464 & 81.00 & 1.05 & 1.31 \\
 (computer) & SAE & \textbf{0.0344} & 0.0465 & \textbf{93.00} & 0.30 & 0.44 \\
 & SAE$_{select}$ & 0.0308 & \textbf{0.0420} & 56.00 & \textbf{17.63} & \textbf{21.64} \\
 
  \midrule
  Binary & TIRM & 0.0103 & \textbf{0.0165} & \textbf{93.00} & 1.29 & 1.46 \\
  (religion) & SAE & \textbf{0.0055} & 0.0240 & \textbf{93.00} & 0.41 & 0.51 \\
  & SAE$_{select}$ & 0.0416 & 0.0468 & 86.00 & \textbf{82.29} & \textbf{82.80} \\
  \midrule

  Multi-class & TIRM & 0.0259 & \textbf{0.0392} & \textbf{98.00} & 1.05 & 1.68 \\
  & SAE & \textbf{0.0201} & 0.0434 & 95.00 & 0.31 & 0.62 \\
  & SAE$_{select}$ & 0.0271 & 0.0446 & 75.00 & \textbf{6.04} & \textbf{8.44} \\
  \midrule

  Multi-labeled &  TIRM & 0.0459 & 0.0467 & 7.00 & \textbf{26.94} & \textbf{31.01} \\
  & SAE & 0.0255 & 0.0492 & \textbf{91.00} & 0.45 & 0.97 \\
  & SAE$_{select}$ & \textbf{0.0255} & \textbf{0.0474} & 90.00 & 0.67 & 1.38 \\
  \bottomrule
  \end{tabular}}
\caption{ Complementary table for \cref{tab:pipeline-comparison-20ng} and \cref{tab:pipeline-eurlex}. All settings keep the same except that CEM uses cut-points at 0 and at 80\% quantile of positive values.}
\label{tab:pipeline-comparison-m128-cutoff0.8}
\end{table*}

\onecolumn
 
\begin{center}
{\Large\bfseries Neuron Interpretations: 20NG Computer}
\end{center}
\vspace{0.5em}
 
\begin{longtable}{@{} >{\raggedright\arraybackslash}p{1.5cm}
                     >{\raggedright\arraybackslash}p{11cm}
                     >{\centering\arraybackslash}p{2cm} @{}}

\toprule
\textbf{Neuron} & \textbf{Interpretation} & \textbf{\# Sim.} \\
\midrule
\endfirsthead
 
\multicolumn{3}{c}{\tablename\ \thetable{} -- continued from previous page} \\
\toprule
\textbf{Neuron} & \textbf{Interpretation} & \textbf{\# Sim.} \\
\midrule
\endhead
 
\midrule
\multicolumn{3}{r}{\textit{Continued on next page}} \\
\endfoot
 
\bottomrule
\endlastfoot
 
0 & Discusses X11 display/graphics color depth—specifically 24-bit/TrueColor visuals or 8-bit colormaps (often mentioning tools like xdpyinfo, xli/xloadimage, XPutImage/XDrawLine, or XGA/X server) & 100 \\
\midrule
7 & Mentions X Window System/OpenWindows/Motif tools or components (e.g., xwd, xterm, xinit, xdm, Xlib, Xt, X11R5) & 100 \\
\midrule
10 & Mentions bitmap image/graphics file formats (e.g., GIF, TIFF, BMP, PCX) in the context of finding or using graphics/conversion/viewer software via FTP/shareware/freeware & 100 \\
\midrule
18 & discussion of 1990s PC hardware bus standards (e.g., VLB/VL-bus, EISA, ISA, MCA, NuBus) and related motherboard/chipset/DMA performance details & 100 \\
\midrule
3 & Classified-style for-sale listing that includes an asking price in dollars (e.g., “\$150”, “Asking \$200”, “sell \$70 only”) for an item being sold & 98 \\
\midrule
9 & Contains the phrase "same problem" (case-insensitive) & 97 \\
\midrule
13 & contains MS-DOS era disk/storage terminology such as FDISK/FORMAT/CMOS partitions and IDE/SCSI hard drives & 93 \\
\midrule
24 & discussion of the Clipper Chip/key escrow (mentions 'Clipper' or 'key escrow' or 'Skipjack') & 93 \\
\midrule
5 & contains electronics/DIY circuit-building content with specific component part numbers or values (e.g., 555, 741, 565, MC14536B, VAC/VDC, ohms, Hz, µF) & 86 \\
\midrule
14 & Discussion of automobiles or motorcycles using specific vehicle makes/models and performance/repair terminology (e.g., Mustang, Miata, Porsche 911, turbo/V-8, transaxle, carbs, clutch, hp) & 85 \\
\midrule
22 & Discussion of Candida/yeast (fungal) infections, often in the context of antibiotics and probiotics (e.g., Lactobacillus, yogurt) & 79 \\
\midrule
25 & Contains explicit Bible scripture citations using book names with chapter:verse notation (e.g., “John 1:1”, “Matthew 28:19”, “Hebrews 10:24-25”). & 79 \\
\midrule
15 & classified for-sale listings that include an explicit asking price marked with a dollar sign (\$) & 75 \\
\midrule
21 & discussion of professional baseball (MLB) teams/standings/predictions (e.g., AL/NL divisions, Yankees/Orioles/Blue Jays, ROY/ROTY, Cy Young) & 75 \\
\midrule
12 & Contains the substring "==clip==" indicating clipped/truncated quoted sections & 70 \\
\midrule
20 & Mentions the Waco siege/Branch Davidians (e.g., Koresh, ATF/BATF, FBI) in a political argument context & 69 \\
\midrule
6 & Discussion of spaceflight/space exploration missions or orbital mechanics (e.g., Moon missions, Earth orbit, planetary probes like Galileo/Cassini/Voyager) & 66 \\
\midrule
11 & Contains discussion of Middle East/Turkey/Armenia/Israel-Palestine political or ethnic conflict (e.g., Arabs/Jews/Zionism/Palestine/Turkiye/Greeks/Armenians) & 60 \\
\midrule
2 & contains political commentary about U.S. government/politicians (e.g., Clinton, Gore, liberals/conservatives) and civil liberties/rights & 58 \\
\midrule
4 & Contains discussion of Major League Baseball players/teams using multiple athlete proper names (e.g., Mays, Mattingly, Winfield) and baseball stats/terms (BA, OBP, steals, HR) & 58 \\
\midrule
30 & contains explicit discussion of U.S. law/constitution/civil-rights enforcement (e.g., mentions the Constitution, amendments, U.S.C., courts, or federal/state criminal charges) & 57 \\
\midrule
17 & discussion of morality/ethics using words like "moral", "morality", "immoral", or "objective morality & 55 \\
\midrule
23 & Contains explicit sarcasm/flame-style Usenet banter with direct insults or profanity (e.g., 'newbie', 'bozo', 'fuck off', 'panties in a bunch') in a quoted-reply thread format & 54 \\
\midrule
48 & Contains the word "Mac" or "Macintosh" (referring to Apple Macintosh computers) & 52 \\
\midrule
99 & Discussion of Apple Macintosh hardware models/upgrades (e.g., Mac IIvx/LC/Centris/SE-30/Duo/PowerBook/Mac Portable) including RAM/memory, accelerators, hard drives, or peripherals & 47 \\
\midrule
31 & contains the literal header field "Archive-name:" (or "Archive-Name:") at the start of an FAQ/archive posting & 45 \\
\midrule
29 & contains explicit deletion/removal marker text such as 'deleted' or '(Deletion)' & 44 \\
\midrule
90 & Contains a quoted/attributed signature line with an identifier like "--" followed by a handle/ID number (e.g., "DoD \#1224", "DoD\#1919") & 44 \\
\midrule
8 & contains direct second-person accusations/challenges using 'you/your' (e.g., 'you have yet to answer', 'you are going to', 'your lack of') & 42 \\

\end{longtable}
\onecolumn
 
\begin{center}
{\Large\bfseries Neuron Interpretations: 20NG Religion}
\end{center}
\vspace{0.5em}
 
\begin{longtable}{@{} >{\raggedright\arraybackslash}p{1.5cm}
                     >{\raggedright\arraybackslash}p{11cm}
                     >{\centering\arraybackslash}p{2cm} @{}}
                     
\toprule
\textbf{Neuron} & \textbf{Interpretation} & \textbf{\# Sim.} \\
\midrule
\endfirsthead
 
\multicolumn{3}{c}{\tablename\ \thetable{} -- continued from previous page} \\
\toprule
\textbf{Neuron} & \textbf{Interpretation} & \textbf{\# Sim.} \\
\midrule
\endhead
 
\midrule
\multicolumn{3}{r}{\textit{Continued on next page}} \\
\endfoot
 
\bottomrule
\endlastfoot
 
25 & Contains explicit Bible scripture citations with book names and chapter:verse notation (e.g., "John 1:1", "Matthew 28:19", "Ephesians 4:4-6") & 100 \\
\midrule
17 & contains frequent discussion of morality/ethics using the words “moral”, “morality”, or “immoral” & 47 \\
 
\end{longtable}
\onecolumn
 
\begin{center}
{\Large\bfseries Neuron Interpretations: 20NG Multi-class (Computer and Religion)}
\end{center}
\vspace{0.5em}
 
\begin{longtable}{@{} >{\raggedright\arraybackslash}p{1.5cm}
                     >{\raggedright\arraybackslash}p{11cm}
                     >{\centering\arraybackslash}p{2cm} @{}}
                     
\toprule
\textbf{Neuron} & \textbf{Interpretation} & \textbf{\# Sim.} \\
\midrule
\endfirsthead
 
\multicolumn{3}{c}{\tablename\ \thetable{} -- continued from previous page} \\
\toprule
\textbf{Neuron} & \textbf{Interpretation} & \textbf{\# Sim.} \\
\midrule
\endhead
 
\midrule
\multicolumn{3}{r}{\textit{Continued on next page}} \\
\endfoot
 
\bottomrule
\endlastfoot
 
25 & Contains explicit Bible scripture references in the form of a book name plus chapter:verse (e.g., 'John 1:1', 'Matthew 5:17', 'Ephesians 4:4-6') & 100 \\
\midrule
18 & Mentions PC motherboard/bus hardware standards (e.g., EISA, VLB/VL-bus, ISA, NuBus, MCA) in a technical discussion & 99 \\
\midrule
7 & Mentions X Window System/OpenWindows/Motif-related tools or components (e.g., xwd, xterm, xdm, xinit, XView, X11R5) & 95 \\
\midrule
10 & Requests or advice about obtaining graphics/image file conversion or viewing software via anonymous FTP, often mentioning formats like GIF/TIFF/BMP/PCX and .ZIP/.EXE filenames & 91 \\
\midrule
14 & discussion of automobiles/motorcycles and their performance or mechanical parts (e.g., car models, engines, transmissions, brakes, horsepower) & 89 \\
\midrule
22 & discussion of Candida/yeast (fungal) infections, often in the context of antibiotic use and probiotics (e.g., Lactobacillus/yogurt) & 89 \\
\midrule
21 & discussion of professional sports (especially MLB baseball) including team names, standings/predictions, or game commentary & 89 \\
\midrule
17 & contains multiple occurrences of the word "moral"/"morality" (including "moral system" or "moral behavior") & 86 \\
\midrule
24 & discussion of the Clipper Chip/key escrow telephone encryption proposal (mentions “Clipper”/wiretap chip/escrowed keys/NSA/NIST/Skipjack/DES in that context) & 86 \\
\midrule
0 & Discusses X11/X Window System display/graphics topics (e.g., XGA, X server visuals, xdpyinfo, xloadimage/xli, pixmaps/colormaps) & 83 \\
\midrule
20 & Mentions the Waco siege/Branch Davidians (e.g., Koresh, ATF/BATF, FBI) in a political discussion & 80 \\
\midrule
11 & contains explicit discussion of Middle East/Turkish/Armenian/Israeli/Palestinian ethnic-national conflict (e.g., mentions Jews/Arabs/Israel/Palestine/Turkiye/Greeks/Armenians/Zionism) & 77 \\
\midrule
3 & classified-style for-sale posting listing items and asking price/offer (e.g., 'for sale/forsale', 'asking \$', 'make an offer', 'buyer pays shipping') & 76 \\
\midrule
5 & contains electronics/DIY circuit-building content with specific component names or part numbers (e.g., 555 timer, 741 op-amp, 565 PLL, MC14536B, MAX641) and voltage/frequency values & 76 \\
\midrule
9 & first-person technical troubleshooting report describing a recurring computer/printer hardware error/problem (often phrased as 'I have the same problem' or asking if others have it) & 71 \\
\midrule
6 & Discussion of space exploration/space missions (e.g., NASA, orbit, Moon, planets, probes like Galileo/Cassini/Voyager, spacecraft/launch systems) & 69 \\
\midrule
4 & Discussion of Major League Baseball players/teams using baseball-specific stats/terms (e.g., batting average/OBP/OPS, steals, pinch-hit ABs, Hall of Fame), with multiple player names & 67 \\
\midrule
2 & Political rant/opinion text about U.S. government/parties/media (explicit mentions like Clinton/Gore/liberals/conservatives/rights/privacy/gun control), rather than technical/marketplace/medical topics & 66 \\
\midrule
23 & Contains emoticon-style smiley faces made with punctuation (e.g., ':-)' or ';)' ) & 64 \\
\midrule
19 & Discussion of ice hockey (NHL/Stanley Cup) including player or team names & 62 \\
\midrule
31 & contains the literal header field "Archive-name:" (Usenet FAQ/archive-style header) & 62 \\
\midrule
12 & Contains the token "==clip==" indicating an edited/truncated quote & 59 \\
\midrule
30 & contains explicit discussion of U.S. law/constitutional rights (e.g., mentions Constitution, amendments, U.S.C., federal courts, or civil rights statutes) & 59 \\
\midrule
15 & Classified for-sale listings that include explicit prices using a dollar sign (\$) for items & 58 \\
\midrule
29 & A standalone deletion marker (the word 'deleted' or '(Deletion)' / 'stuff deleted...') indicating removed content & 56 \\
\midrule
8 & contains multiple rhetorical questions marked with '?' & 55 \\
\midrule
13 & mentions DOS-era PC disk/drive management terms like FDISK/format/partitions and IDE vs SCSI hard drives & 52 \\
\midrule
27 & discussion of RS-232/serial COM port hardware or wiring (e.g., RS232, serial port, COM1/COM3, null modem, pinout, baud/parity) & 49 \\
\midrule
26 & Text about riding motorcycles (e.g., bikes, riding, helmets/gear, passenger/pillion advice, specific motorcycle models like Ninja/CBR/GSX-R) & 47 \\
\midrule
73 & contains the substring “LD” (uppercase L followed by uppercase D) somewhere in the text & 46 \\
\midrule
58 & contains at least one question mark (“?”) in the text & 43 \\
\midrule
28 & Contains a signature delimiter line consisting solely of two hyphens ("--") on its own line & 43 \\
\midrule
93 & discussion of Hell in a Christian/Biblical context (explicit mentions of Hell plus Bible passages/books like Luke, Matthew, Revelation, or doctrines like Atonement/resurrection) & 43 \\
\midrule
55 & Mentions MSG (monosodium glutamate) explicitly & 41 \\
\midrule
48 & Mentions Apple Macintosh computers/products (e.g., Mac, Macintosh, Mac SE/Mac II, Apple, QuickDraw) & 40 \\
 
\end{longtable}
\onecolumn
 
\begin{center}
{\Large\bfseries Neuron Interpretations}
\end{center}
\vspace{0.5em}
 
\begin{longtable}{@{} >{\raggedright\arraybackslash}p{1.5cm}
                     >{\raggedright\arraybackslash}p{11cm}
                     >{\centering\arraybackslash}p{2cm} @{}}
                     
\toprule
\textbf{Neuron} & \textbf{Interpretation} & \textbf{\# Sim.} \\
\midrule
\endfirsthead
 
\multicolumn{3}{c}{\tablename\ \thetable{} -- continued from previous page} \\
\toprule
\textbf{Neuron} & \textbf{Interpretation} & \textbf{\# Sim.} \\
\midrule
\endhead
 
\midrule
\multicolumn{3}{r}{\textit{Continued on next page}} \\
\endfoot
 
\bottomrule
\endlastfoot
 
2 & EU legal acts specifically about veterinary/animal health measures (mentions terms like 'veterinary', 'animal health', and/or diseases such as foot-and-mouth disease, scrapie, BSE/TSE) & 100 \\
\midrule
13 & EU legal acts that are Directives (titled "DIRECTIVE" or "COUNCIL/COMMISSION DIRECTIVE" with an /EC or /EEC number) specifically about wheeled agricultural or forestry tractors & 100 \\
\midrule
14 & EU legal act text that explicitly refers to Spain/Spanish appointments (contains the word 'Spanish' or 'Spain' in the title/body) & 100 \\
\midrule
18 & Mentions Central/Eastern European (post-communist) countries such as Bulgaria, Romania, Czechoslovakia/Czech Republic, Slovakia/Slovak, Hungary, Poland, Slovenia, Yugoslavia/FYROM, or former Soviet Union republics & 100 \\
\midrule
21 & Contains the exact phrase "establishing the standard import values for determining the entry price of certain fruit and vegetables & 100 \\
\midrule
31 & Commission Regulation establishing the standard import values for determining the entry price of certain fruit and vegetables & 100 \\
\midrule
101 & Contains the exact phrase 'Only the French text is authentic' & 100 \\
\midrule
4 & contains the phrase "Community support framework" (often in the title "on the establishment/approving the Community support framework for Community structural assistance") & 99 \\
\midrule
9 & Council Decision appointing member(s) and/or alternate member(s) of the Committee of the Regions & 99 \\
\midrule
15 & EU legal acts specifically about the marketing/sowing/authorization of agricultural seed (mentions “seed” in the context of arable crops/plant varieties, e.g., cereal/maize/vegetable seed) & 99 \\
\midrule
44 & Text is a Commission Regulation establishing “unit values for the determination of the customs value of certain perishable goods” (repeated phrase in the title/body) & 99 \\
\midrule
5 & Contains the exact phrase "concerning the classification of certain goods in the Combined Nomenclature & 98 \\
\midrule
17 & Begins with the phrase “COUNCIL DECISION” (including the variant “COUNCIL AND COMMISSION DECISION”) rather than “COMMISSION REGULATION/DECISION” & 98 \\
\midrule
22 & Contains the phrase 'protected designations of origin and protected geographical indications' (often with PDO/PGI) in the title/opening lines & 98 \\
\midrule
24 & Contains the phrase "standing invitation to tender" (often with "special/individual invitation to tender") & 98 \\
\midrule
25 & Text is a Commission Regulation amending Council Regulation (EC) No 881/2002 on restrictive measures linked to Usama bin Laden, the Al‑Qaida network and the Taliban (often phrased as 'amending for the Nth time') & 98 \\
\midrule
89 & Contains the exact phrase "private storage aid" (in the context of EU regulations granting private storage aid for cheeses) & 98 \\
\midrule
0 & Document header begins with five asterisks: "*****" (often "***** COMMISSION REGULATION (EEC)"/"***** COMMISSION DECISION") & 97 \\
\midrule
27 & Commission Regulation (EC) text fixing the maximum export refund on wholly milled round grain rice (often tied to an invitation to tender) & 97 \\
\midrule
6 & Issued by the European Parliament and the Council (contains the header phrase “THE EUROPEAN PARLIAMENT AND THE COUNCIL OF THE EUROPEAN UNION”) & 95 \\
\midrule
7 & Contains the exact phrase "representative prices and additional duties" (in the context of amending them for imports in the sugar sector) & 95 \\
\midrule
16 & Document header begins with the phrase “COUNCIL IMPLEMENTING” (as in “COUNCIL IMPLEMENTING REGULATION …” or “COUNCIL IMPLEMENTING DECISION …”). & 95 \\
\midrule
26 & EU Commission regulations specifically about butter intervention tenders, using the phrase 'fixing the maximum purchasing/buying-in price for butter' or 'fixing the minimum selling prices for butter' (often also mentioning aid for cream/butter/concentrated butter) under Regulation (EC) No 2571/97 or 2771/1999 & 95 \\
\midrule
33 & Contains the exact phrase "establishing unit values for the determination of the customs value of certain perishable goods & 95 \\
\midrule
65 & Mentions the 'Canary Islands' in the title/opening text (often as part of 'forecast supply balance' regulations under Council Regulation (EEC) No 1601/92) & 95 \\

\end{longtable}

\end{document}